\documentclass[lettersize,journal]{IEEEtran}
\usepackage{amsmath,amsfonts}
\usepackage{algorithmic}
\usepackage{algorithm}
\usepackage{array}
\usepackage{textcomp}
\usepackage{stfloats}
\usepackage{url}
\usepackage{verbatim}
\usepackage{graphicx}
\usepackage{cite}
\usepackage{wrapfig}
\usepackage[utf8]{inputenc} 
\usepackage{booktabs}       
\usepackage{amsfonts}       
\usepackage{epsfig}
\usepackage{amsmath}
\usepackage{amssymb}
\usepackage{stmaryrd}
\usepackage{dsfont}

\usepackage{amsthm}
\usepackage{subcaption}
\theoremstyle{definition}
\newtheorem{definition}{Definition}[section]
\theoremstyle{remark}

\usepackage{afterpackage}
\usepackage{hyperref}
\hypersetup{
    colorlinks=True,
    citecolor=blue,
    }
\usepackage{url}

\usepackage{eso-pic}
\usepackage{xcolor}

\begin{document}

\title{Zero-Shot Neural Network Evaluation with Sample-Wise Activation Patterns}

\author{Yameng Peng,
Andy Song,
Haytham M. Fayek,
Vic Ciesielski,
Xiaojun Chang
\thanks{Yameng Peng, Andy Song, Haytham M. Fayek and Vic Ciesielski are with School of Computing Technologies, RMIT University, Australia. Xiaojun Chang is with Department of Electronic Engineering and Information Science, University of Science and Technology of China. Email: 1024peng@gmail.com; haytham.fayek@ieee.org; xjchang@ustc.edu.cn; \{andy.song,vic.ciesielski\}@rmit.edu.au;}
}

\markboth{IEEE TRANSACTIONS ON PATTERN ANALYSIS AND MACHINE INTELLIGENCE}%
{}



\newcommand{\FirstPageIEEEFooter}{%
\AddToShipoutPictureFG*{%
  \AtPageLowerLeft{%
    \raisebox{18pt}{%
      \makebox[\paperwidth]{%
        \parbox{0.94\paperwidth}{%
          \centering\footnotesize
          \copyright~2026 IEEE. All rights reserved, including rights for text and data mining
          and training of artificial intelligence and similar technologies. Personal use is permitted,
          but republication/redistribution requires IEEE permission. See
          https://www.ieee.org/publications/rights/index.html for more information.\\
          DOI: 10.1109/TPAMI.2026.3691075
        }%
      }%
    }%
  }%
}%
}

\FirstPageIEEEFooter

\maketitle

\begin{abstract}
Zero-shot proxies, also known as training-free metrics, are widely adopted to reduce the computational overhead in neural network evaluation for scenarios such as Neural Architecture Search (NAS), as they do not require any training. Existing zero-shot metrics have several limitations, including weak correlation with the true performance and poor generalisation across different networks or downstream tasks. For example, most of these metrics apply only to either convolutional neural networks (CNNs) or Transformers, but not both. To address these limitations, we propose Sample-Wise Activation Patterns (SWAP), and its derivative, SWAP-Score, a novel and highly effective zero-shot metric. 
SWAP-Score is broadly applicable across both architecture families and task domains, demonstrating strong predictive performance in the majority of tasks. This metric measures the expressivity of neural networks over a mini-batch of samples, showing a high correlation with the neural networks' ground-truth performance. For both CNNs and Transformers, the SWAP-Score outperforms existing zero-shot metrics across computer vision and natural language processing tasks. For instance, Spearman's correlation coefficient between the SWAP-Score and CIFAR-10 validation accuracy for DARTS CNNs is 0.93, and 0.71 for FlexiBERT Transformers on GLUE tasks. Moreover, SWAP-Score is label-independent, hence can be applied at the pre-training stage of language models to estimate their performance for downstream tasks.  When applied to NAS, SWAP-empowered NAS, SWAP-NAS can achieve competitive performance using only approximately 6 and 9 minutes of GPU time, on CIFAR-10 and ImageNet respectively. Our code is available at: \url{https://github.com/pym1024/SWAP_Universal}.
\end{abstract}

\begin{IEEEkeywords}
Zero-shot, Training-free, Performance Evaluation, Convolutional Neural Network, Transformer, Neural Architecture Search. 
\end{IEEEkeywords}

\section{Introduction}
 \label{intro}
\IEEEPARstart{P}{erformance} evaluation of neural networks is essential in the deep learning field, spanning areas such as knowledge distillation, reinforcement learning, neural architecture search, and model pruning \cite{Ref:135,Ref:136,Ref:70,Ref:88,Ref:77,Ref:137}. Conventional approaches evaluate neural network performance via back-propagation training, which often leads to prohibitively high computational costs in certain research areas. In knowledge distillation, training a student model to replicate the teacher model’s behaviour through back-propagation imposes considerable computational demands, especially with large teacher models \cite{Ref:138,Ref:139}. Similarly, AlphaGO achieved superhuman performance in the game of Go through reinforcement learning and Monte Carlo Tree Search, but this required thousands of self-play games, each with complex evaluations and back-propagation steps \cite{Ref:140}. Neural architecture search (NAS), which aims to automatically design high-performing neural networks for specific tasks, requires substantial computational resources due to the need to evaluate numerous candidate networks \cite{Ref:01,Ref:101,Ref:102}. To alleviate these costs, various alternatives have been proposed, including performance predictors, early-stopping, and weight-sharing strategies \cite{Ref:27,Ref:41,Ref:103,Ref:72,Ref:17,Ref:10}.

An emerging approach is the utilisation of zero-shot metrics, also known as training-free or zero-cost proxies \cite{Ref:71,Ref:92,Ref:105,Ref:70,Ref:89,Ref:91,Ref:119}. These metrics aim to eliminate the need for network training entirely as they exhibit either positive or negative correlations with the networks' ground-truth performance. Typically, these metrics necessitate only a few forward or backward passes with a mini-batch of input data, rendering their computational costs negligible compared to traditional network performance evaluation methods. However, zero-shot metrics face several challenges, including unreliable correlation with the network's ground-truth performance \cite{Ref:71,Ref:89} and limited generalisation across different types of neural networks and downstream tasks.  In some cases, these metrics struggle to consistently outperform simple counterparts such as model parameters or FLOPs, which are computationally low-cost \cite{Ref:100}.
\begin{figure}[t!]
  \begin{center}
    \includegraphics[width=0.9\linewidth, height=0.32\textheight]{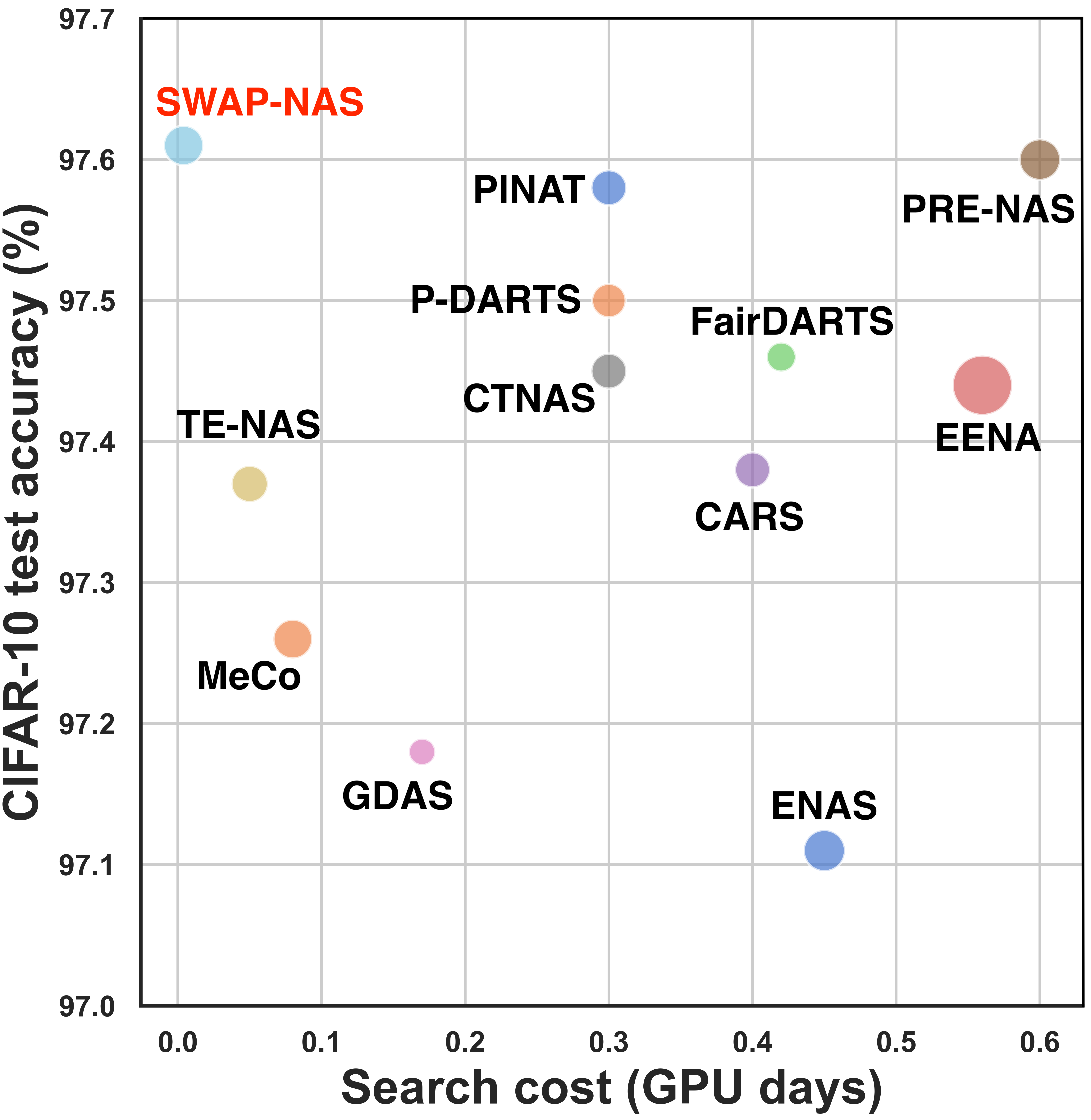}
  \end{center}
  \vspace{-3mm}
    \caption{Search cost and performance on CIFAR-10 for SWAP-NAS compared to state-of-the-art NAS methods. Only methods requiring less than 1 GPU day are shown. The dot size indicates model size, highlighting that SWAP-NAS attains superior performance with significantly lower search cost.}
  \label{methods_compare}
\end{figure}

\IEEEpubidadjcol
To overcome these limitations, we introduce Sample-Wise Activation Patterns (SWAP), and its derivative, SWAP-Score, a novel, broadly applicable, and highly effective zero-shot metric. The inspiration is from studies on network expressivity \cite{Ref:77,Ref:78}, whilst addressing the aforementioned limitations.

SWAP-Score generalises well on neural networks based on piecewise linear and non-linear activation functions, exhibiting strong correlations with the networks' ground-truth performance across various task types. This study rigorously evaluates its predictive capabilities on ReLU-based convolutional networks derived from five distinct architecture spaces — NAS-Bench-101, NAS-Bench-201, NAS-Bench-301, TransNAS-Bench-101-Micro/Macro \cite{Ref:44,Ref:45,Ref:53,Ref:117} — across eight different computer vision tasks. SWAP-Score is benchmarked against 15 existing training-free metrics for the correlation between metrics and networks' ground-truth performance. Furthermore, similar comparisons are performed on GELU-based Transformers from the FlexiBERT architecture space \cite{Ref:123} with the General Language Understanding Evaluation (GLUE) tasks \cite{Ref:124}. Finally, we demonstrate SWAP-Score's capability by integrating it into NAS as SWAP-NAS. This method combines the efficiency of SWAP-Score with the effectiveness of population-based evolutionary search, which is typically computationally intensive. The \textbf{primary contributions} of this work are as follows:

\begin{itemize}
\item  We introduce a novel, broadly applicable, and highly effective zero-shot metric, SWAP-Score, based on \textbf{S}ample-\textbf{W}ise \textbf{A}ctivation \textbf{P}atterns. SWAP-Score demonstrates significantly higher capability in evaluating networks compared to state-of-the-art (SOTA) zero-shot metrics.  Its robust, superior performance, and high generalisation capabilities are validated through comprehensive experiments across various types of neural networks, including convolutional neural networks (CNNs) and Transformers, spanning a range of computer vision and natural language processing tasks.
\item We apply SWAP-Score to a practical scenario—neural architecture search (NAS)—which traditionally requires extensive network training. By integrating SWAP-Score with an evolutionary algorithm, we enable a highly efficient NAS algorithm, SWAP-NAS, capable of completing a search on CIFAR-10 in just 0.004 GPU days (6 minutes), outperforming state-of-the-art NAS methods in both speed and performance, as illustrated in Fig. \ref{methods_compare}. A direct search on ImageNet takes only 0.006 GPU days (9 minutes) to achieve SOTA accuracies, highlighting its exceptional efficiency and performance.
\end{itemize}

\section{Related Work}
\label{related_work}
In deep learning, full evaluation by back-propagation training is the standard approach for assessing a neural network's performance on specific tasks. Neural Architecture Search (NAS) \cite{Ref:01} is a prominent example where dense evaluations are required. For instance, AmoebaNet \cite{Ref:08} employs an evolutionary search strategy that trains each sampled network from scratch, consuming roughly 3150 GPU days on CIFAR-10.

To mitigate this cost, subsequent studies explored few-shot or early-stop evaluation \cite{Ref:12,Ref:24,Ref:26}, where architectures are only trained for a small number of epochs to approximate final accuracy. Although this reduces training time, the correlation between early and final accuracy can be unreliable, especially across diverse architecture families.

A further step involves proxy evaluation, where networks are trained on simplified tasks to serve as a proxy for full evaluation \cite{Ref:39,Ref:40}. Examples include training on a subset of the dataset, using reduced image resolutions, or restricting model depth/width. These proxies lower cost but often fail to generalise across tasks or search spaces, and the accuracy of ranking remains limited.

To address these limitations, performance predictors were introduced. These methods fit regression models on architecture–accuracy pairs to predict unseen architectures’ performance \cite{Ref:40, Ref:69, Ref:103, Ref:31, Ref:41}. While effective once trained, this strategy requires preparing the predictor dataset itself, which demands substantial computational overhead.

Another widely used approach is weight sharing, where candidate architectures inherit parameters from a supernet rather than being trained independently \cite{Ref:15, Ref:35, Ref:64, Ref:24, Ref:10, Ref:17, Ref:30}. For example, DARTS \cite{Ref:10} combines a one-shot supernet with gradient-based optimisation, requiring only 4 GPU days to reach 97.33\% test accuracy on CIFAR-10. However, parameter sharing across heterogeneous networks is problematic, and supernet optimisation is complex and difficult to generalise to broader settings \cite{Ref:31, Ref:104}.

More recently, zero-shot metrics have emerged, eliminating the need for training altogether \cite{Ref:71, Ref:92, Ref:105, Ref:70, Ref:89, Ref:91,Ref:119,Ref:141}. For example, TE-NAS \cite{Ref:71} combines the number of linear regions \cite{Ref:78} with the spectrum of the neural tangent kernel \cite{Ref:79}, reducing the search cost to 0.05 GPU days on CIFAR-10 and 0.17 on ImageNet. NWOT \cite{Ref:70} measures the overlap of activations between input samples in untrained networks, while Zen-Score \cite{Ref:92} estimates network expressivity by averaging Gaussian complexity from random inputs. With a specialised search space, Zen-NAS achieves 83.6\% top-1 accuracy on ImageNet within 0.5 GPU day, marking the first zero-shot NAS to outperform training-based methods. MeCo \cite{Ref:141} was proposed as a correlation-based proxy. It leverages the minimum eigenvalue of the Pearson correlation matrix across feature maps, and eliminates the need for back-propagation and data labels.
Alongside CNN-oriented metrics, Transformer-specific indicators have also been introduced. TF-NAS proposed the DSS-indicator, combining synaptic diversity for self-attention with synaptic saliency for MLP modules \cite{Ref:133}, and NAS-BERT-Benchmark introduced Attention Confidence, measuring the certainty of each attention head \cite{Ref:122}.

However, an empirical study highlights the limitations of existing training-free metrics. NAS-Bench-Suite-Zero \cite{Ref:100} evaluated 13 such methods across multiple tasks and reported that most do not generalise well across architecture families and tasks. In some cases, simple baselines such as model parameter count or FLOPs even outperform more complex metrics, motivating the need for new approaches with stronger cross-domain generalisation.

\section{Sample-Wise Activation Patterns and SWAP-Score}
\label{tf}
To address the above challenges, we introduce Sample-Wise Activation Patterns and its derivative, SWAP-Score. Similar to NWOT and Zen-Score, SWAP-Score draws inspiration from studies on network expressivity, aiming to uncover the expressivity of deep neural networks by examining their activation patterns. What sets SWAP-Score apart from other training-free metrics is its foundation on sample-wise activation patterns, which offers several benefits including high correlation with the networks' ground-truth performance and good generalisation across different types of neural networks and tasks.

The following subsections are organised as: Section \ref{num_lr} introduces the concept of standard activation patterns, which can reveal a network's expressivity. Section \ref{swap}, sample-wise activation patterns and SWAP-Score are presented.  Section \ref{reg_func} introduces a regularisation approach to SWAP-Score, specifically designed for NAS scenarios.

\subsection{Standard Activation Patterns, Network's Expressivity \& Limitations }
\label{num_lr}
Studies exploring the expressivity of deep neural networks \cite{Ref:77,Ref:106,Ref:78} demonstrate that networks employing piecewise linear activation functions, such as ReLU \cite{Ref:114}, partition their input space into two regions: either zero or a non-zero positive value. These ReLU activation functions introduce piecewise linearity into the network. Since the composition of piecewise linear functions remains piecewise linear, a ReLU neural network can be viewed as a piecewise linear function. Consequently, the input space of such a network can be divided into multiple distinct segments, each referred to as a linear region. The number of distinct linear regions serves as an indicator of the network's functional complexity.  A network with more linear regions is capable of capturing more complex features in the data, thereby exhibiting higher expressivity. Although the concept of linear region is originally defined for piecewise linear activation functions, we extend this idea to piecewise non-linear activation functions like GELU \cite{Ref:125}.

\begin{definition} 
Given $\mathcal{N}$ as a deep neural network with piecewise linear activation function, $\theta$ as a fixed set of network parameters (randomly initialised weights and biases) of $\mathcal{N}$, a batch of inputs containing $S$ samples, the standard activation pattern, $\mathbb{A}_{\mathcal{N},\theta}$, is defined as a set of post-activation values shown as follows:
\end{definition}

\begin{equation}
\label{lr_set}
\mathbb{A}_{\mathcal{N},\theta} = \left\{ \mathbf{p}^{(s)} : \mathbf{p}^{(s)} = \mathds{1}(p_v^{(s)})_{v=1}^{V},~ s \in \{1, \ldots, S\} \right\},
\end{equation}

\noindent where $V$ denotes the number of intermediate values feeding into activation functions. $p_v^{(s)}$ denotes a single post-activation value from the $v^{th}$ intermediate value at $s^{th}$ sample. $\mathds{1}(x)$ is the indicator function. We adopt the $Signum$ function as the indicator function as shown in Eq. \ref{sign}.
\begin{equation}
\label{sign}
\mathds{1}(x) = 
\begin{cases} 
-1 & \text{if } x < 0 \\
0 & \text{if } x = 0 \\
1 & \text{if } x > 0
\end{cases}
\end{equation}
For ReLU-based neural networks, the indicator function converts positive non-zero values to one whilst leaving zero values unchanged. For GELU-based neural networks, the indicator function converts positive non-zero values to one, negative values to negative one and leaves zero values unchanged. Consequently, $\mathbb{A}_{\mathcal{N},\theta}$ represents a set containing the binarised or ternarised post-activation values produced by network $\mathcal{N}$ with parameters $\theta$ and $S$ input samples. 
This design choice is motivated by both efficiency and robustness. Continuous activation magnitudes vary substantially across different architectures. Thresholding with the Signum function provides a compact and noise-resistant representation, while also enabling efficient counting of distinct activation patterns at scale.

Following the concept discussed in \cite{Ref:77,Ref:106,Ref:78}, the network's expressivity can be revealed by counting the cardinality of a set composed of activation patterns.

\begin{figure}[!h]
  \begin{center}
    \includegraphics[width=0.9\linewidth,height=0.15\textheight]{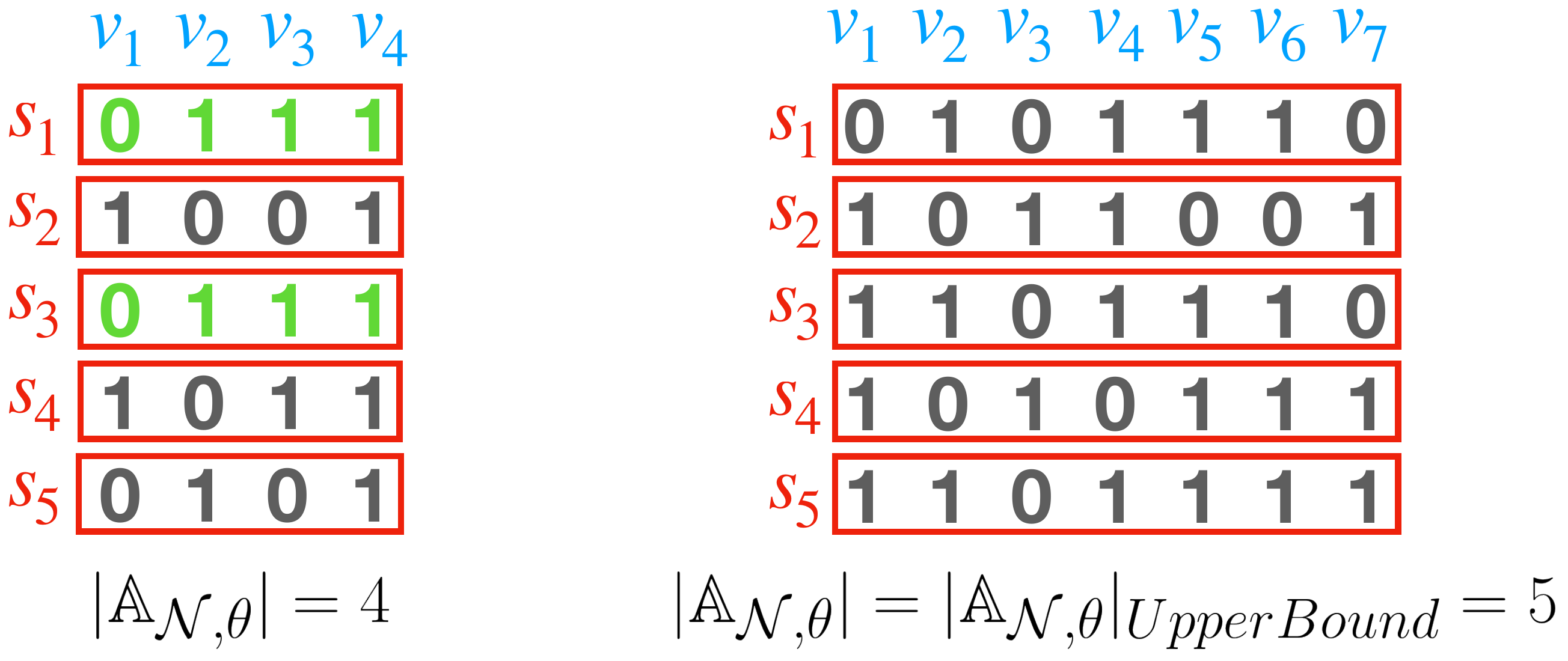}
  \end{center}
  \caption{Examples of activation pattern matrices $\mathbb{A}_{\mathcal{N},\theta}$ for the same network with different inputs. Duplicate activation patterns are highlighted in green.}
  \label{lr_compare_1}
\end{figure}

\textbf{Limitation.} As illustrated in Fig. \ref{lr_compare_1}, the set $\mathbb{A}_{\mathcal{N},\theta}$ can be viewed as a matrix, with each element as a row representing a vector $\mathds{1}(p_v^{(s)})_{v=1}^{V}$ of binarised or ternarised post-activation values over all intermediate values in $V$.  Each value or cell corresponds to $\mathds{1}(p_v^{(s)})$ as defined in Eq. \ref{lr_set}. Since $V$ represents the number of intermediate values feeding into the activation layers, define $\mathcal{N}$ contains $L$ layers, we have:

\begin{equation}
\label{inter_value_eq}
\begin{cases} 
V_{\text{CNN}} = \sum_{l=1}^{L} \left( c_l \times (\left\lfloor \frac{{w_l - k_l}}{t_l} \right\rfloor + 1) \times (\left\lfloor \frac{{h_l - k_l}}{t_l} \right\rfloor + 1) \right), \\ \\
V_{\text{Transformer}} = \sum_{l=1}^{L} T \times d_{\text{ff}}.
\end{cases}    
\end{equation}

For CNNs, $c_l$ denotes the number of convolution kernels, $t_l$ denotes the stride of convolution kernels, $k_l$ denotes the kernel size, $w_l$ and $h_l$ are the width and height of the feature map in layer in $l^{th}$ layer. For Transformers, $T$ denotes the sequence length, $d_{ff}$ is the dimension of the feed-forward network's hidden layer. Hence, the length of $\mathds{1}(p_v^{(s)})_{v=1}^{V}$ is set by the dimensionality of the input samples and depth of the network. 

The upper bound of the cardinality of $\mathbb{A}_{\mathcal{N}}$ is equal to the number of input samples, $S$. Given the same number of inputs, higher-dimensional inputs or deeper networks will generate more intermediate values, i.e., $V \gg S$, making it more likely to produce distinct vectors and reach the upper bounds of cardinality. As a simplified illustration, Fig. \ref{lr_compare_1} (left) shows the matrix $\mathbb{A}_{\mathcal{N},\theta}$ derived from low-dimensional inputs, while that of Fig. \ref{lr_compare_1} (right) is derived from higher-dimensional inputs.  Due to pattern duplication, the cardinality in the former is $4$, whereas in the latter is $5$.  Note, the latter reaches the upper limit, the number of input samples, $S$, that is 5 in this case. This highlights the limitation of examining standard activation patterns for measuring the network's expressivity.  Methods leverage standard activation patterns typically require inputs of small dimensions, e.g., $1 \times 3 \times 3$ \cite{Ref:71}. Otherwise, the metric values from different networks will all approach the number of input samples, $S$, making them indistinguishable.

\subsection{Sample-Wise Activation Patterns}
\label{swap}
Sample-Wise Activation Patterns address the limitation identified above. It preserves the information within the activation values, while bringing a significantly higher upper bound for the cardinality. This gives its derivative, SWAP-Score, a better capability to distinguish networks with different performance in a more subtle way.
 
\begin{definition}[Sample-Wise Activation Patterns]
Given a deep neural network $\mathcal{N}$, $\theta$ as a fixed set of network parameters (randomly initialised weights and biases) of $\mathcal{N}$, a batch of inputs containing $S$ samples, sample-wise activation patterns $\mathbb{\hat{A}}_{\mathcal{N},\theta}$ is defined as follows:

\begin{equation}
\label{swap_set}
\mathbb{\hat{A}}_{\mathcal{N},\theta} = \left\{ \mathbf{p}^{(v)} : \mathbf{p}^{(v)} = \mathds{1}(p_s^{(v)})_{s=1}^{S},~ v \in \{1, \ldots, V\} \right\},
\end{equation}

\noindent where $p_{s}^{(v)}$ denotes a single post-activation value from the $s^{th}$ sample at the $v^{th}$ intermediate value.
\end{definition}
Note, in comparison with $\mathbb{A}_{\mathcal{N},\theta}$ in Eq. \ref{lr_set}, the vectors here are now sample-wise (over neurons) rather than value-wise (over samples) as in standard activation patterns. In sample-wise activation patterns, $\mathds{1}(p_s^{(v)})_{s=1}^{S}$ is a vector containing binarised or ternarised post-activation values across all samples in $S$. 

\begin{definition}[SWAP-Score $\Psi$]
Given a SWAP set $\mathbb{\hat{A}}_{\mathcal{N},\theta}$, the SWAP-Score $\Psi$ of network $\mathcal{N}$ with a fixed set of network parameters $\theta$ is defined as the cardinality of the set, computed as follows: 
 
\begin{equation}
\label{swap_eq}
\mathbf{\Psi}_{\mathcal{N},\theta} = \bigg\vert \mathbb{\hat{A}}_{\mathcal{N},\theta} \bigg\vert .
\end{equation}
\end{definition}

\begin{figure}[!h]
  \begin{center}
    \includegraphics[width=0.9\linewidth,height=0.17\textheight]{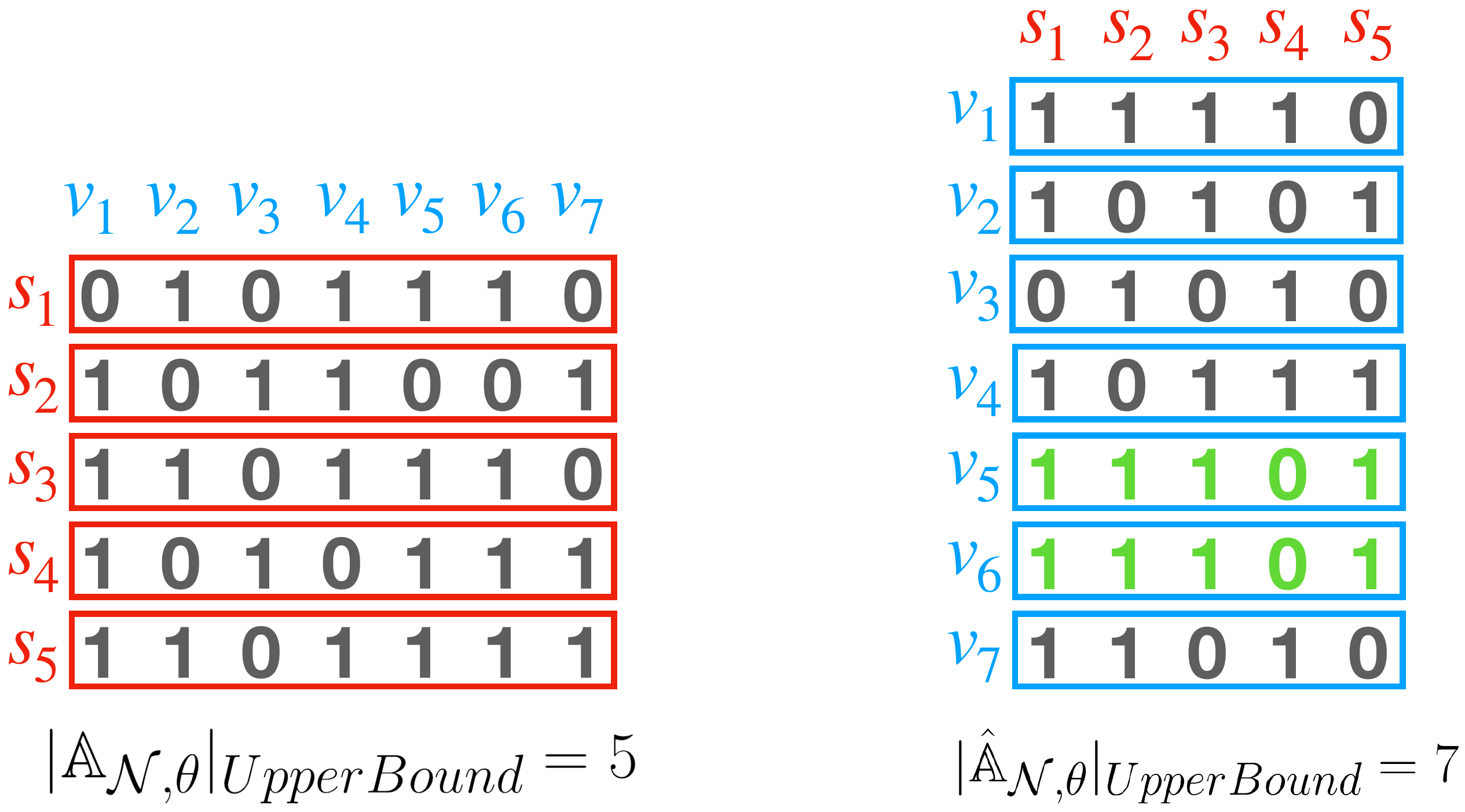}
  \end{center}
  \caption{Comparison between original activation pattern matrix $\mathbb{A}_{\mathcal{N},\theta}$ and sample-wise activation pattern matrix $\mathbb{\hat{A}}_{\mathcal{N},\theta}$ for the same network. Duplicate patterns are denoted in green. Sample-wise activation pattern matrix offer a higher capacity for unique patterns.}
  \label{lr_demo}
\end{figure}

Fig. \ref{lr_demo} illustrates the connection and the difference between $\mathbb{A}_{\mathcal{N},\theta}$ and $\mathbb{\hat{A}}_{\mathcal{N},\theta}$ in a simplified form.  Both sets are based on the same network with the same input. Hence they have the same set of binarised or ternarised post-activation values but are represented differently.  The upper bound of the cardinality using standard activation patterns, $\mathbb{A}_{\mathcal{N},\theta}$, is $5$. In contrast, the upper bound of sample-wise activation patterns, $\mathbb{\hat{A}}_{\mathcal{N},\theta}$, is extended to $7$. According to Eq. \ref{inter_value_eq}, the number of intermediate values $V$ grows exponentially with either an increase in the dimensionality of the input samples or the depth of the neural networks. This implies that the number of intermediate values, $V$, would be much larger than the number of input samples, $S$. As a result, SWAP has a significantly higher capacity for distinct patterns, which allows SWAP-Score to measure the network's expressivity more fine-grained. Specifically, this characteristic leads to a high correlation with the ground-truth performance of network $\mathcal{N}$ (see Section \ref{experiment} for more details).

\subsection{Regularisation}
\label{reg_func}
In the NAS scenarios, zero-shot metrics have a well-known drawback.  They tend to favour larger models \cite{Ref:102}, meaning they do not naturally lead to smaller models when such models are desirable. Using a convolutional neural network as an example, the convolution operations with larger kernel sizes or more channels have more parameters while producing more intermediate values compared to operations like skip connection \cite{Ref:04} or pooling layer. Consequently, larger networks typically yield higher metric values, which may not always be desirable. To mitigate this bias, we add regularisation to SWAP-Score.

\begin{definition}[Regularisation]
Given the total number of network parameters, $\Theta$ , coefficients, $\mu$ and $\sigma$, SWAP regularisation is defined as follows:
\begin{equation}
\label{reg_factor}
f(\Theta) = e^{-(\frac{(\Theta - \mu)^2}{\sigma})}.
\end{equation}  
\end{definition}

\begin{definition}[Regularised SWAP-Score]
Given regularisation function $f(\Theta) $, SWAP-Score $\Psi$ of network $\mathcal{N}$ with a fixed set of network parameters $\theta$, regularised SWAP-Score ${\Psi}^{\prime}$ is defined as: 
\begin{equation}
\label{reg_swap_eq}
\mathbf{\Psi}^{\prime}_{\mathcal{N},\theta} = \mathbf{\Psi}_{\mathcal{N},\theta} \times f(\Theta).
\end{equation}
\end{definition}

Regularisation function $f(\Theta)$ is a bell-shaped curve. Coefficient $\mu$ controls the centre position of this curve. Coefficient $\sigma$ adjusts the width of the curve. By explicitly setting the values for $\mu$ and $\sigma$, the regularised SWAP-Score, $\mathbf{\Psi}^{\prime}_{\mathcal{N},\theta}$, can guide the resulting architectures toward a desired range of model sizes.  

\section{Experiments}
\label{experiment}
Comprehensive experiments are conducted to confirm the effectiveness and generalisation of SWAP-Score. For ReLU-based networks and computer vision tasks, SWAP-Score is benchmarked against 16 other zero-shot metrics across five distinct search spaces and seven tasks (Section \ref{sec_metrics_compare_cv}). For GELU-based networks and natural language processing tasks, SWAP-Score is compared with 10 zero-shot metrics, including some used in computer vision experiments and those specifically designed for Transformers and NLP tasks, on 500 BERT-like Transformers from the FlexiBERT \cite{Ref:123}. Subsequently, the regularised SWAP-Score is integrated with evolutionary search as SWAP-NAS to evaluate its performance in a practical scenario, Neural Architecture Search (NAS), which traditionally involves a large amount of network training. Furthermore, SWAP-NAS is compared with SOTA NAS methods in terms of both search performance and efficiency (Section \ref{search_darts}).

\subsection{SWAP-Score on ReLU-based Networks and Computer Vision Tasks}
\label{sec_metrics_compare_cv}

\begin{figure*}[!hb]
  \begin{center}
    \includegraphics[width=\linewidth,height=0.4\textheight]{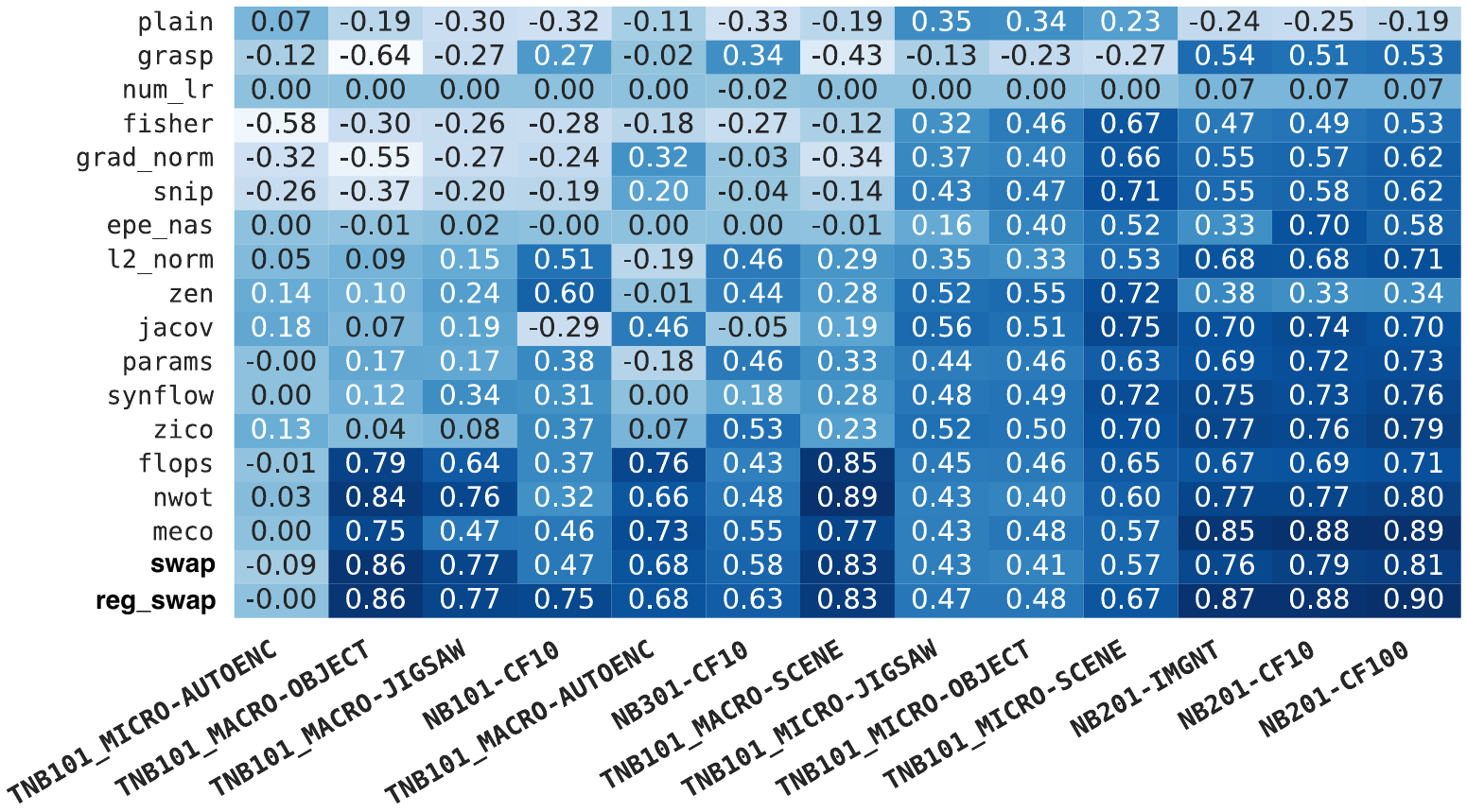}
  \end{center}
    \caption{Spearman's correlation coefficients between 18 training-free metrics (rows), including SWAP-Score and its regularised variant, across task - search space pairs (columns), Each cell reports Spearman's correlation coefficient between metric value and the ground-truth performance, computed over 1000 architectures randomly sampled from the given space and averaged over five independent runs. Rows and columns are ordered by mean values. SWAP-Score and its regularised variant consistently rank among the top-performing metrics across most tasks. The colour scale denotes coefficients in $[-1, 1].$}
  \label{cv_metrics_compare}
\end{figure*}

For computer vision tasks, the networks are trained in a conventional supervised manner, which involves directly training networks on the training set until convergence and validating on the test set. The computer vision tasks are:
\begin{enumerate}
    \item \textbf{Image classification tasks}: CIFAR-10 / CIFAR-100 \cite{Ref:85}, ImageNet-1k \cite{Ref:86} and ImageNet16-120 \cite{Ref:99}.
    \item \textbf{Object detection task}: Taskonomy dataset \cite{Ref:115}.
    \item \textbf{Scene classification task}: MIT Places dataset \cite{Ref:116}.
    \item \textbf{Jigsaw puzzle}: the input is divided into patches and shuffled according to preset permutations. The objective is to classify which permutation is used \cite{Ref:100}.
    \item \textbf{Autoencoding}: a pixel-level prediction task that encodes images into low-dimensional latent representation then reconstructs the raw image \cite{Ref:100}.
\end{enumerate}

Five architecture spaces are used to verify the effectiveness of SWAP-Score on ReLU-based networks. These include micro/cell-based search spaces, which focus on searching for optimal building blocks or cells, and macro search spaces, which focus on searching for the entire network architecture. The specific spaces are:
\begin{enumerate}
    \item \textbf{NAS-Bench-101} \cite{Ref:44}: a cell-based benchmark search space which contains 423624 unique architectures trained on CIFAR-10. The architectures are designed as ResNet-like and Inception-like \cite{Ref:04,Ref:107}.
    \item \textbf{NAS-Bench-201} \cite{Ref:45}: a cell-based benchmark search space which contains 15625 unique architectures trained on CIFAR-10, CIFAR-100 and ImageNet16-120.
    \item \textbf{NAS-Bench-301} \cite{Ref:53}: a surrogate benchmark space which contains architectures sampled from the DARTS search space.
    \item \textbf{TransNAS-Bench-101-Mirco/Macro} \cite{Ref:117}: consisting of a micro (cell-based) search space of size 4096, and a macro (stack-based) search space of size 3256. 
\end{enumerate}

SWAP-Score is compared against 15 zero-shot metrics \cite{Ref:70,Ref:88,Ref:92,Ref:105,Ref:108,Ref:109,Ref:110,Ref:111,Ref:112,Ref:119} across different architecture spaces and tasks, in terms of correlation.  Our regularised SWAP-Score is also included in the comparison as it is used in the NAS experiments on computer vision tasks that will be presented later. The extensive experiments follow exactly the same setup as NAS-Bench-Suite-Zero \cite{Ref:100}, which is a standardised framework for verifying the effectiveness of zero-shot metrics. The hyper-parameters, such as batch size, input data, sampled architectures and random seeds are fixed and consistently applied to all these zero-shot metrics as in NAS-Bench-Suite-Zero. Specifically, the metrics' values are calculated by feeding a mini-batch of samples from each target dataset, then calculating the correlation between metrics' values and the network's ground-truth performance of each target dataset. 

\begin{figure*}[!hb]
  \begin{center}
    \includegraphics[width=\linewidth,height=0.5\textheight]{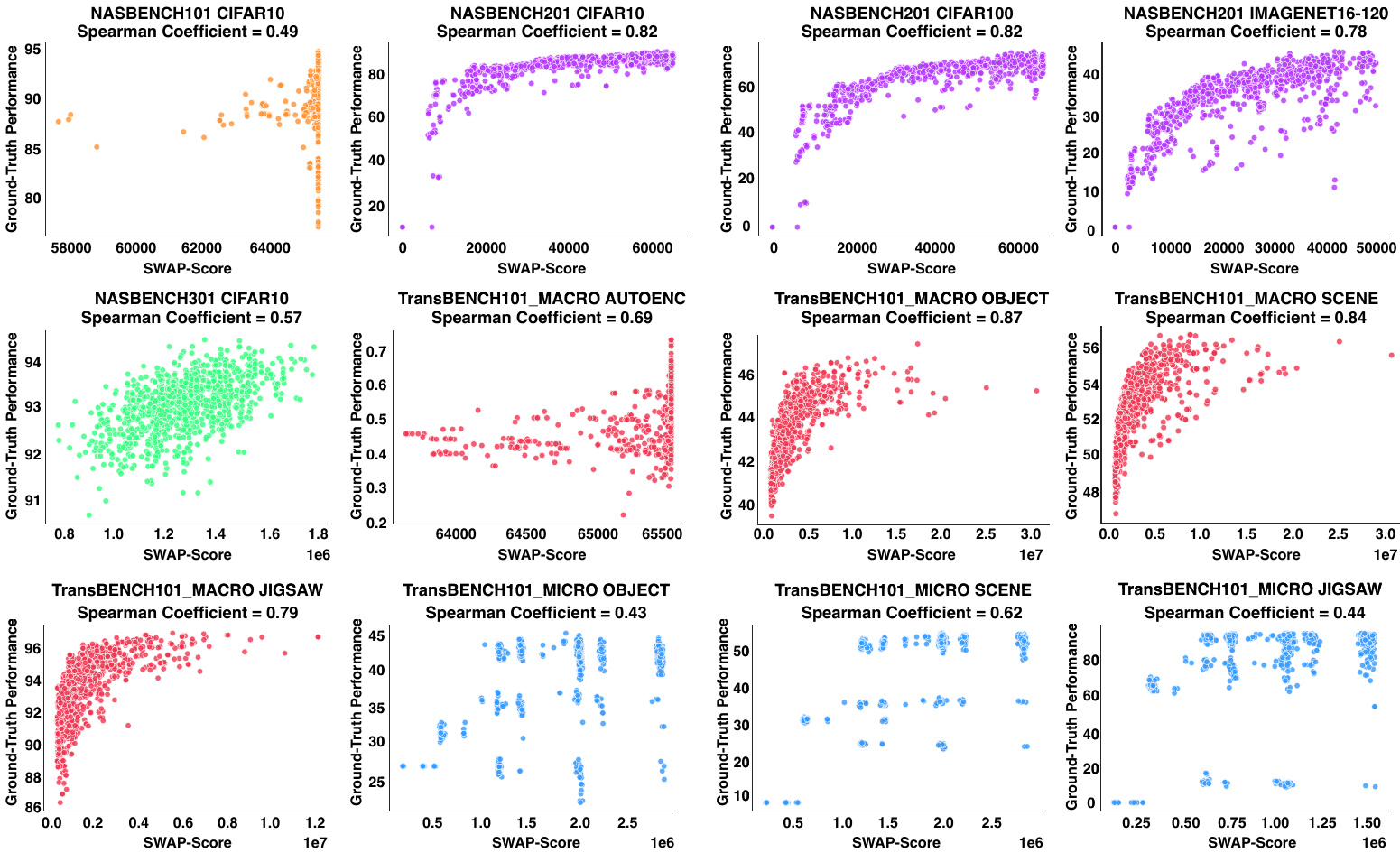}
  \end{center}
   \caption{Scatter plots of SWAP-Score vs. ground-truth performance for architectures from NAS-Bench-101/201/301 and TransNAS-Bench-101.  Each subplot corresponds to a vision dataset, and each dot represents a single architecture. The strong correlations confirm that SWAP-Score reliably predicts performance across diverse search spaces.}
  \label{swap_cv}
\end{figure*}

The comparison is shown in Fig. \ref{cv_metrics_compare}, where each column represents the Spearman coefficients of all metrics on one task with one given search space. They are computed on 1000 randomly sampled architectures. Each value in Fig. \ref{cv_metrics_compare} is an average of five independent runs with different random seeds. For regularised SWAP-Score, the $\mu$ and $\sigma$ factors for the regularisation function are determined by the model size distribution based on 1000 randomly sampled architectures. This process requires only a few seconds for each search space.

The results in Fig. \ref{cv_metrics_compare} clearly demonstrate the superior predictive capability of SWAP-Scores across diverse types of architectural spaces and tasks. Notably, both SWAP-Score and its regularised version outperform 15 other metrics in the majority of the evaluations. An interesting observation is a significant enhancement in SWAP-Score's performance when regularisation is applied (noted as 'reg\_swap'), although its original intention is to control the model size during architecture search. This improvement is particularly evident in cell-based search spaces, including NAS-Bench-101, NAS-Bench-201, NAS-Bench-301, and TransNAS-Bench-101-Micro. However, it is worth noting that regularisation does not appear to impact the correlation results in the macro architecture space, TransNAS-Bench-101-Macro.

\begin{figure*}[!ht]
  \begin{center}
    \includegraphics[width=\linewidth,height=0.5\textheight]{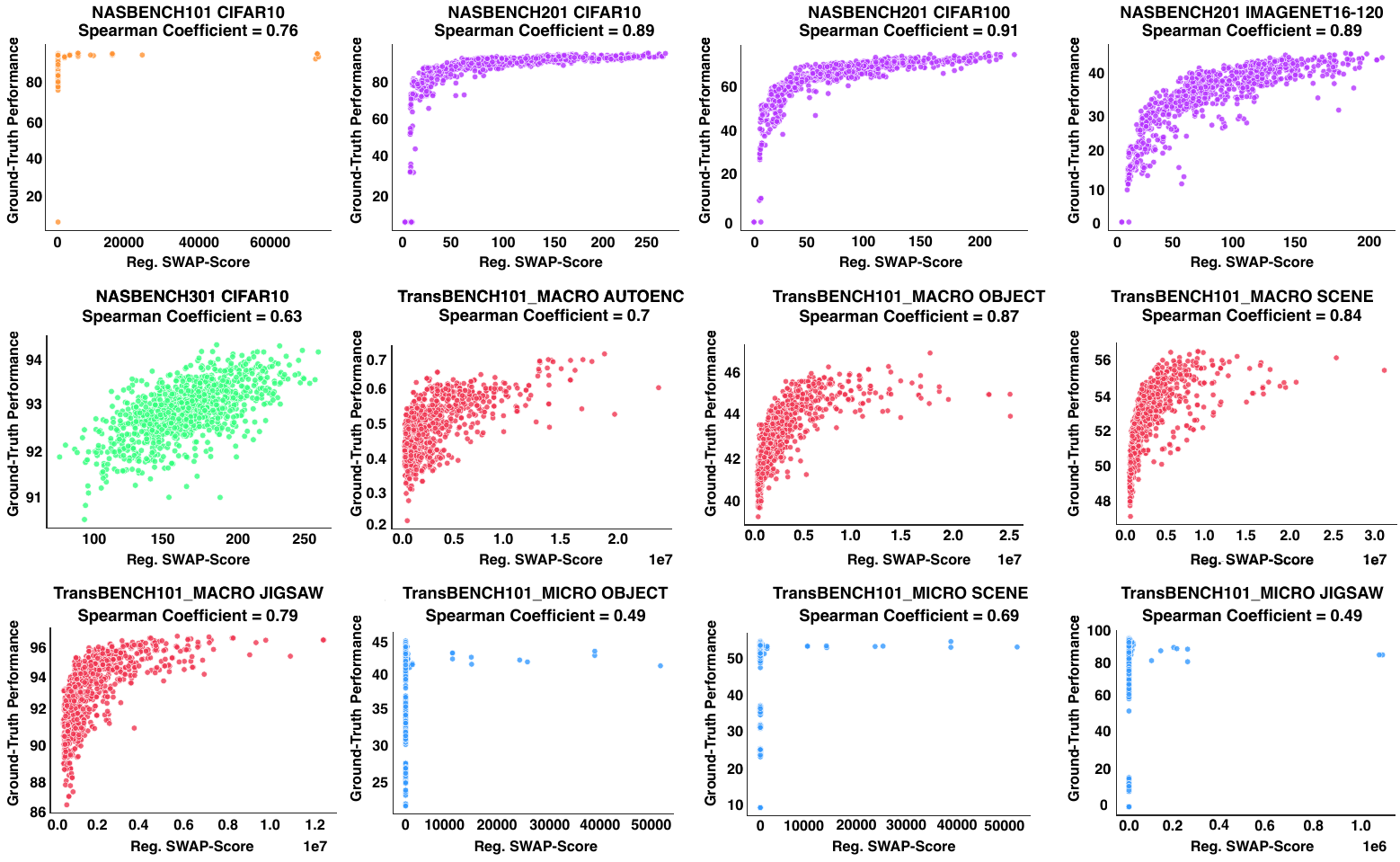}
  \end{center}
  \caption{Scatter plots of regularised SWAP-Score vs. ground-truth performance for architectures from NAS-Bench-101/201/301 and TransNAS-Bench-101. Each subplot corresponds to a vision dataset, and each dot represents a single architecture. Regularised SWAP-Score maintains or improves correlation with true accuracy, showing robustness to variance across tasks.}
  \label{reg_swap_cv}
\end{figure*}

Figures \ref{swap_cv} and \ref{reg_swap_cv} illustrate the detailed relationships between SWAP-Score, its regularised version, and networks' ground-truth performance across each architecture space and computer vision task presented in Fig. \ref{cv_metrics_compare}. Each dot in these figures indicates a distinct ReLU-based network. From the figures, it can be observed that the regularisation makes dots more concentrated in the micro architectural spaces, i.e., NAS-Bench-101/201/301, thus leading to a higher correlation. However, in the macro architecture spaces, the distribution of the dots remains almost the same. The hypothesis is that final networks from micro architecture spaces are constructed by repeating the same building blocks/cells multiple times, thus their performance is more linear with the number of model parameters. Hence, when regularisation is applied based on model parameters to micro networks, the performance aligns more closely with the final models.

\subsection{SWAP-Score on GELU-based Networks and Natural Language Processing Tasks}
\label{sec_metrics_compare_nlp}

For natural language processing tasks, following the setup in NAS-BERT-Benchmark \cite{Ref:122}, we leverage 500 unique GELU-based Transformer networks which sampled from the FlexiBERT architecture space \cite{Ref:123} and pre-trained on the OpenWebText corpus \cite{Ref:126}, which contains approximately 40GB of text data from webpages. Specifically, these networks follow ELECTRA-Small setup and pre-training scheme \cite{Ref:134}. Subsequently, these 500 pre-trained Transformers are fine-tuned and evaluated on 8 NLP tasks from the General Language Understanding Evaluation (GLUE) benchmark \cite{Ref:124}. The final GLUE score is calculated as a weighted average of the individual task scores, for each Transformer network. Specifically, the natural language processing tasks are:
\begin{enumerate}
    \item \textbf{Corpus of Linguistic Acceptability (CoLA)}: Single-sentence classification, determines whether a sentence is grammatically acceptable \cite{Ref:127}.
    \item \textbf{Multi-Genre Natural Language Inference(MNLI)}: Multi-class classification, determines whether a hypothesis entails, contradicts, or is neutral to a premise \cite{Ref:128}.
    \item \textbf{Microsoft Research Paraphrase Corpus (MRPC)}: Sentence pair classification, identifies whether two sentences are paraphrases \cite{Ref:129}.
    \item \textbf{Question Natural Language Inference (QNLI)}: Sentence pair classification, determines whether a sentence contains the answer to a question \cite{Ref:130}.
    \item \textbf{Quora Question Pairs (QQP)}: Sentence pair classification, determines whether two questions are semantically equivalent \cite{Ref:124}.
    \item \textbf{Recognising Textual Entailment (RTE)}: Sentence pair classification which determines whether one text entails another \cite{Ref:124}.
    \item \textbf{Stanford Sentiment Treebank (SST)}: Single-sentence classification, determines the sentiment of a movie review \cite{Ref:131}.
    \item \textbf{Semantic Textual Similarity (STS)}: Regression task, evaluate the semantic similarity of two sentences \cite{Ref:132}.
\end{enumerate}

The metric calculation for natural language processing tasks differs from the experiments in computer vision tasks, as Transformers typically follow a pre-train and fine-tune approach. Thus, the goal is to avoid pre-training and directly predict the networks' performance on downstream tasks. The metrics' values are calculated by feeding a mini-batch of samples from the pre-training dataset (OpenWebText), then calculating the correlation between metrics' values and the network's ground-truth performance on the downstream task (GLUE). No labels are available during the metric calculation. Most training-free metrics designed for ReLU-based CNNs are not applicable for Transformers, as they require either target labels or specific components of CNNs. In contrast, the SWAP-Score is label-independent and only requires activation values, thus it can work smoothly on both CNNs and Transformers.

\begin{table*}[!ht]
\scriptsize
\centering
\caption{Spearman's correlation coefficients between 11 zero-shot metrics (including our SWAP-Score) and GLUE score of 500 BERT-like Transformers.}
\begin{tabular}{c|c|c|c|c|c|c|c|c|c|c|c}
\hline
                                                               & \begin{tabular}[c]{@{}c@{}}Synaptic\\ Diversity\end{tabular} & \begin{tabular}[c]{@{}c@{}}Synaptic\\ Saliency\end{tabular} & \begin{tabular}[c]{@{}c@{}}Activation \\ Distance\end{tabular} & \begin{tabular}[c]{@{}c@{}}Jacobian\\ Score\end{tabular} & \begin{tabular}[c]{@{}c@{}}Attention\\ Confidence\end{tabular} & \begin{tabular}[c]{@{}c@{}}Attention\\ Importance\end{tabular} & \begin{tabular}[c]{@{}c@{}}Attention Softmax\\ Confidence\end{tabular} & DSS-Indicator & \#Params & \#FLOPs & SWAP-Score     \\ \hline
\begin{tabular}[c]{@{}c@{}}Spearman\\ Coefficient\end{tabular} & 0.169                                                        & 0.177                                                       & 0.031                                                          & 0.016                                                    & 0.666                                                          & 0.162                                                          & 0.055                                                                  & -0.071        & 0.653    & 0.652   & \textbf{0.711} \\ \hline
\end{tabular}
\label{nlp_metrics_compare}
\end{table*}

\begin{figure*}[!ht]
  \begin{center}
    \includegraphics[width=\linewidth,height=0.3\textheight]{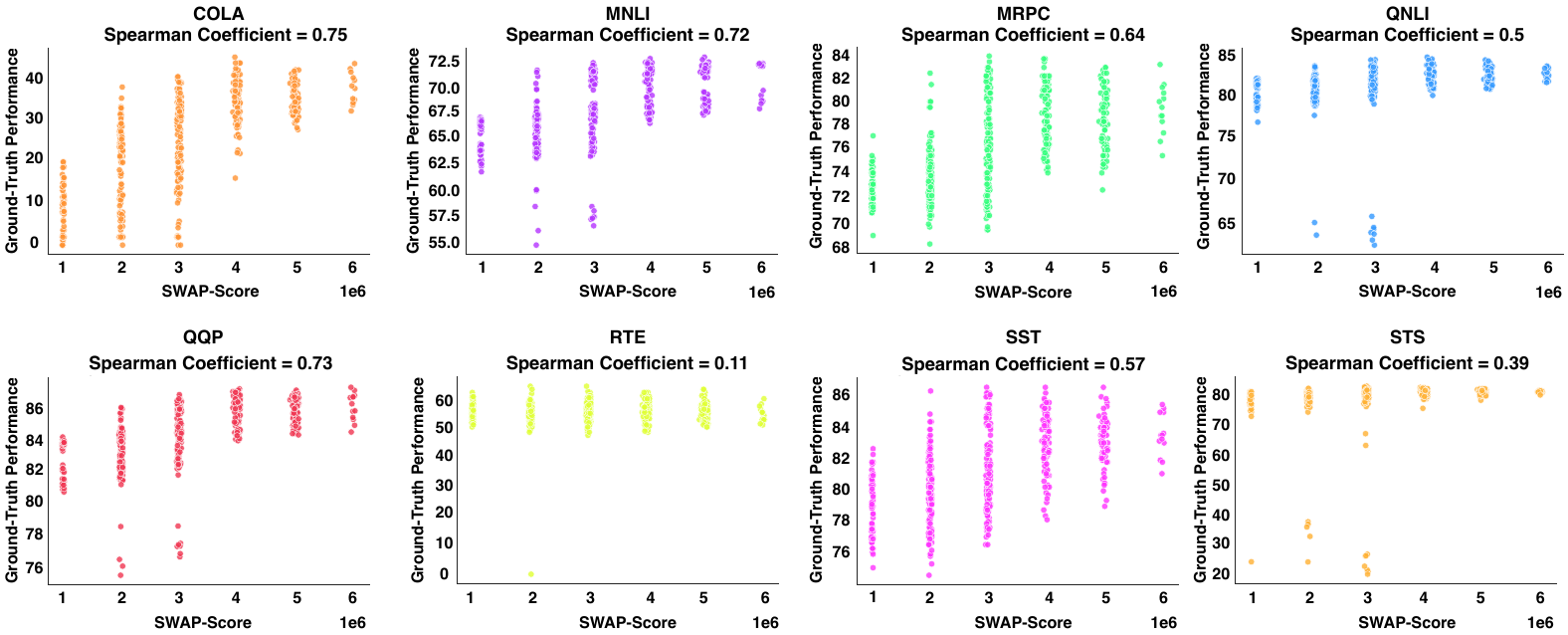}
  \end{center}
    \caption{Scatter plots of SWAP-Score vs. ground-truth performance for Transformer models from FlexiBERT search space on GLUE tasks. Each subplot corresponds to a NLP dataset, and each dot represents a single architecture. Results demonstrate that SWAP-Score generalises beyond CNNs to Transformer architectures, maintaining strong predictive ability on the majority of NLP tasks.}
  \label{swap_nlp}
\end{figure*}

Table \ref{nlp_metrics_compare} shows the comparison between SWAP-Score and 10 other zero-shot proxies. The results demonstrate the superior predictive capability of SWAP-Score on Transformer networks, achieving a Spearman's correlation coefficient of 0.711, outperforming all other existing zero-shot metrics, which are either universal metrics or specially designed for Transformers. Notably, the number of model parameters (\#Params) and model FLOPs (\#FLOPs) outperform most metrics that are more computationally expensive. 

\begin{figure}[!h]
  \begin{center}
    \includegraphics[width=\linewidth,height=0.35\textheight]{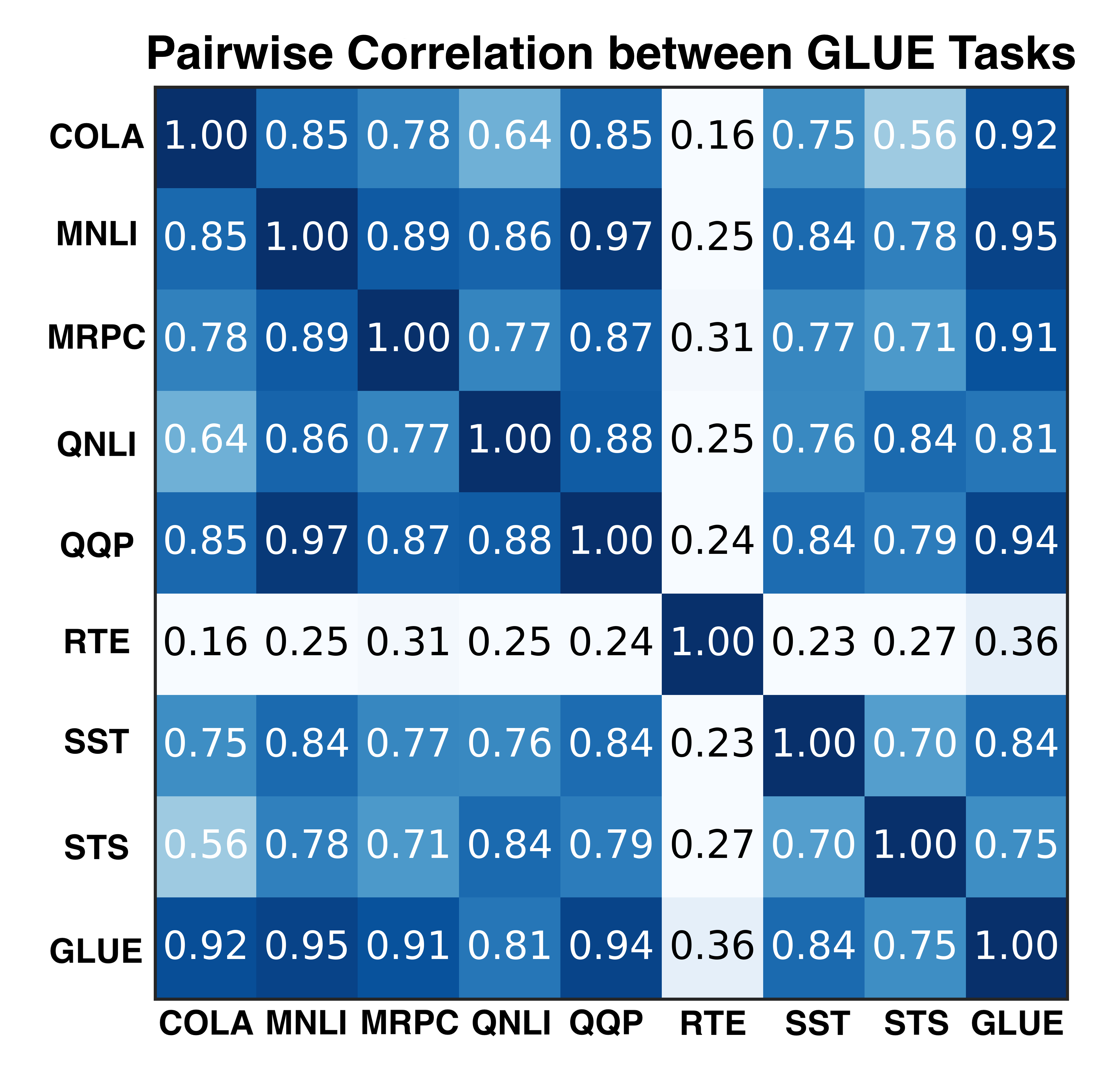}
  \end{center}
      \caption{Pairwise Spearman’s correlation among the ground-truth performance of 500 BERT-like Transformers across GLUE tasks. High correlations indicate that models which perform well on one task tend to perform well on others, revealing transferability and consistency of relative model rankings across NLP tasks. While most tasks exhibit strong correlations, the RTE task shows weak alignment with others, indicating limited transferability.}
  \label{glue_tasks_corr}
\end{figure}
Fig. \ref{swap_nlp} shows the detailed results of SWAP-Score for each task in the GLUE benchmark. SWAP-Score demonstrates strong correlation with the majority of tasks, except for RTE. To analyse this phenomenon, we investigated the performance correlation at the task level, as shown in Fig. \ref{glue_tasks_corr}. The ground-truth performance between most GLUE tasks is strongly correlated, except for RTE. 
The RTE task exhibits a much weaker correlation with other GLUE tasks. This behaviour is not due to the proposed metric itself, but reflects the underlying characteristics of the task \cite{Ref:124}. First, RTE is considerably smaller in scale than most other GLUE datasets, containing only a few thousand labelled examples. This limited size increases variance in model rankings and makes performance less stable across architectures. Second, the linguistic nature of RTE differs from tasks such as SST-2 or MNLI, it involves fine-grained textual entailment over short sentence pairs, which shares less overlap with the features that benefit other tasks. As a result, models that excel on other GLUE tasks do not necessarily achieve consistent gains on RTE \cite{Ref:142}.

\subsection{SWAP-Score for Neural Architecture Search}
\label{search_darts}
To further validate the effectiveness of SWAP-Score in a practical scenario, we utilised it for NAS by integrating the regularised SWAP-Score with evolutionary search, referred to as SWAP-NAS. The DARTS search space was used for the experiments, given its widespread presence in NAS studies, allowing a fair comparison with state-of-the-art methods. For the evolutionary search algorithm, SWAP-NAS is similar to \cite{Ref:08}, but uses the regularised SWAP-Score as the performance measure. Parent architectures generate possible offspring iteratively in each search cycle, employing both mutation and crossover operations. Unlike many training-based NAS approaches that initiate the search on the CIFAR-10 and later transfer the architecture to the ImageNet, SWAP-NAS conducts direct searches on the ImageNet dataset. This is made feasible due to the high efficiency of the SWAP-Score.
 
\subsubsection{Evolutionary Search Algorithm} 

\begin{algorithm}[h!]
\caption{SWAP-NAS}
\label{swap_nas_algo}
\begin{algorithmic}[1]
\REQUIRE Population size $P$, Search cycle $C$, Sample size $S$, $ mutation\_times$, Training-free metric \textit{SWAP}
\STATE $\mathit{population}$ $\gets \emptyset$
\WHILE {$population$  $< P$}
\STATE $\mathit{model.arch} \gets RandomGenerateNetworks()$ 
\STATE $\mathit{model.score} \gets$ \textit{SWAP}$(model.arch)$ 
\STATE Add $model$ to the $population$
\ENDWHILE
\FOR {$c = 1,2,...,C$}
\STATE $\mathit{candidates}\gets$ $S$ random samples of $population$
\STATE $\mathit{parent} \gets$ best in $candidates$ \textbf{OR} best from crossover between the best and the second best in $candidates$
\WHILE {$mutation\_times$ not reached}
\STATE $\mathit{child} \gets$ $Mutate(parent)$
\STATE $\mathit{child.score} \gets$ \textit{SWAP}$(child.arch)$
\ENDWHILE
\STATE Add the best $child$ to the $population$
\STATE Remove the worst from the $population$
\ENDFOR
\end{algorithmic}
\end{algorithm}

\begin{figure}[h!]
  \begin{center}
    \includegraphics[width=0.7\linewidth,height=0.18\textheight]{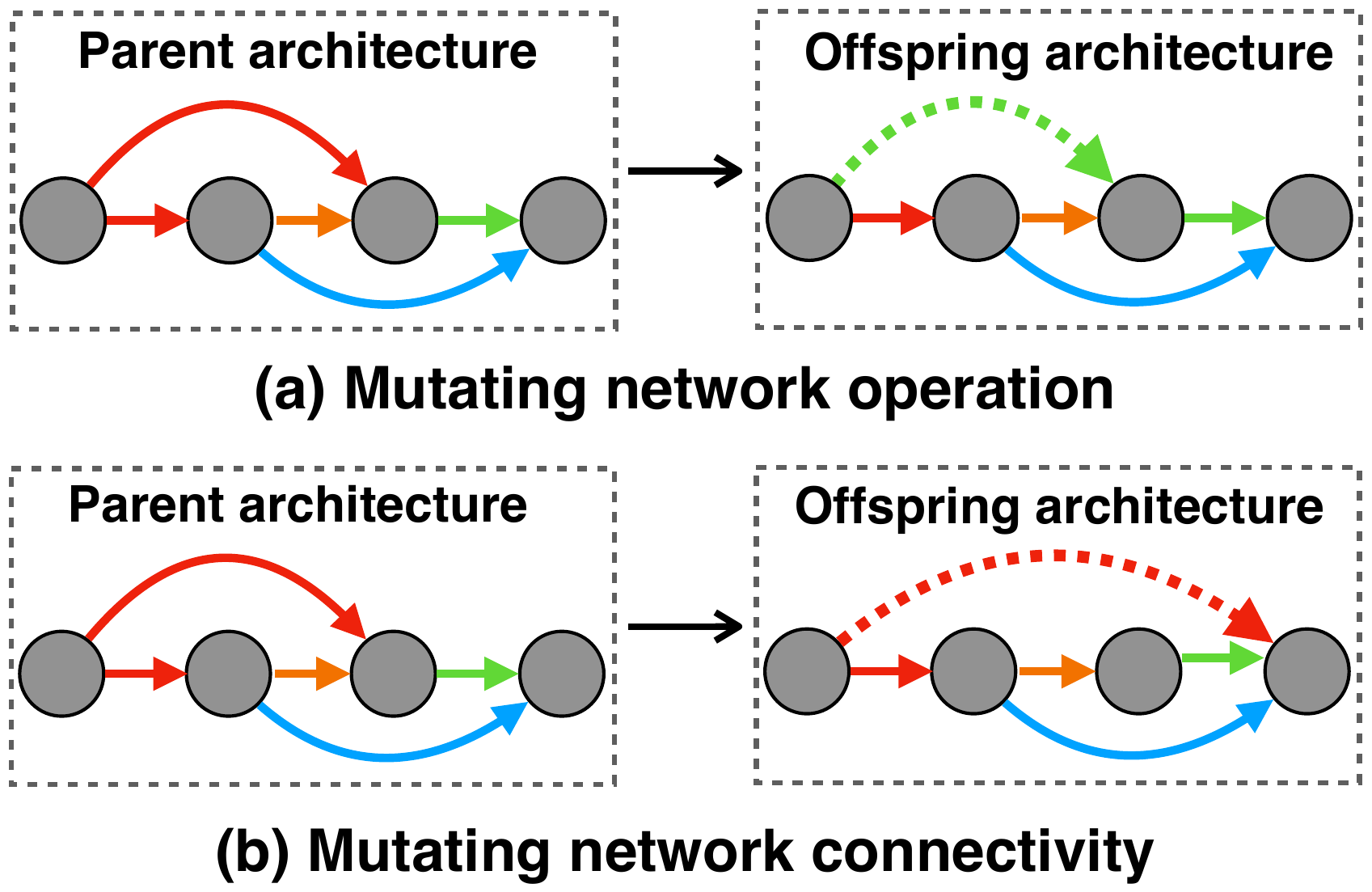}
  \end{center}
    \caption{Illustration of the mutation operators in SWAP-NAS. (a) Operation mutation alters the computational block within the parent network. (b) Connectivity mutation modifies inter-layer connections. Both strategies diversify offspring architectures and improve search coverage.}
  \label{mutations}
\end{figure}

Algorithm \ref{swap_nas_algo} details the steps of SWAP-NAS. SWAP-Score replaces the traditional back-propagation training as the performance measurement for the candidate networks (Steps 4 \& 12). A tournament strategy is used to sample offspring networks, with half of the networks randomly selected from the population in each search cycle (Step 8). Then, the search algorithm randomly decides whether to perform a crossover operation on the top two candidates of the sampled networks or directly use the best network as the parent (Step 9). The offspring networks are generated by continuously applying mutation to the parent (Step 11). There are two types of mutation: network operation mutation and connectivity mutation, as shown in Fig. \ref{mutations}.

\begin{table*}[ht!]
\centering
\caption{Performance comparison between the networks found by SWAP-NAS and other methods on CIFAR-10. A lower test error rate is better. $\dagger$ means the method adopted DARTS space. $\star$ indicates the original paper only reported their best result.}
\begin{tabular}{l|c|c|c|c|c}
\hline
  & \textbf{\begin{tabular}[c]{@{}c@{}}Test Error (\%)\end{tabular}} & \textbf{\begin{tabular}[c]{@{}c@{}}Params (M)\end{tabular}} & \textbf{\begin{tabular}[c]{@{}c@{}}GPU Days\end{tabular}} &\textbf{\begin{tabular}[c]{@{}c@{}}Search Method\end{tabular}} & \begin{tabular}[c]{@{}c@{}}{\textbf{Evaluation}}\end{tabular} \\ \hline
PNAS \cite{Ref:40} & 3.34$\pm$0.09   & 3.2 & 225  & SMBO      & Predictor \\
EcoNAS \cite{Ref:58}$\dagger$ & 2.62$\pm$0.02 & 2.9 & 8 & Evolution & Conventional \\
DARTS \cite{Ref:10}$\dagger$     & 3.00$\pm$0.14 & 3.3 & 4   & Gradient & One-shot \\
EvNAS \cite{Ref:62}$\dagger$ & 2.47$\pm$0.06 & 3.6 & 3.83 & Evolution & One-shot \\
RandomNAS \cite{Ref:32}$\dagger$ & 2.85$\pm$0.08 & 4.3 & 2.7 & Random & One-shot \\
EENA \cite{Ref:51} & 2.56$\star$ & 8.47 & 0.65 & Evolution & Weights Inherit \\
PRE-NAS \cite{Ref:103}$\dagger$  & 2.49$\pm$0.09 & 4.5 & 0.6 & Evolution & Predictor \\
ENAS \cite{Ref:17} & 2.89$\star$ & 4.6 & 0.45 & Reinforce & One-shot  \\ 
FairDARTS \cite{Ref:35}$\dagger$ & 2.54$\star$ & 2.8 & 0.42 & Gradient & One-shot \\
CARS \cite{Ref:07}$\dagger$ & 2.62$\star$ & 3.6 & 0.4 & Evolution & One-shot \\
P-DARTS \cite{Ref:52}$\dagger$   & 2.50$\star$ & 3.4 & 0.3 & Gradient & One-shot \\
TNASP \cite{Ref:121}$\dagger$   & 2.57$\pm$0.04 & 3.6 & 0.3 & Evolution & Predictor\\
PINAT \cite{Ref:120}$\dagger$   & 2.54$\pm$0.08 & 3.6 & 0.3 & Evolution & Predictor \\
GDAS \cite{Ref:63}   & 2.82$\star$ & 2.5 & 0.17 & Gradient & One-shot \\
CTNAS \cite{Ref:72} & 2.59$\pm$0.04 & 3.6 & 0.1+0.3 & Reinforce & Predictor \\
MeCo \cite{Ref:141}$\dagger$ & 2.69$\pm$0.05 & 4.2 & 0.08 & Pruning & Training-free 
\\
TE-NAS \cite{Ref:71}$\dagger$ & 2.63$\pm$0.064 & 3.8 & 0.03 & Pruning & Training-free 
\\
\hline
\textbf{SWAP-NAS-A ($\mu$=0.9, $\sigma$=0.9)} $\dagger$ & 2.65$\pm$0.04 & 3.06 & 0.004 & Evolution & Training-free \\
\textbf{SWAP-NAS-B ($\mu$=1.2, $\sigma$=1.2)} $\dagger$ & 2.54$\pm$0.07 & 3.48 & 0.004 & Evolution & Training-free \\
\textbf{SWAP-NAS-C ($\mu$=1.5, $\sigma$=1.5)} $\dagger$ & 2.48$\pm$0.09 & 4.3 & 0.004 & Evolution & Training-free \\
\textbf{SWAP-NAS (Adaptive)} $\dagger$ & 2.53$\pm$0.07 & 3.51 & 0.004 & Evolution & Training-free \\
\hline
\end{tabular}
\label{cifar10results}
\end{table*}

\subsubsection{Practical choice of $\mu$ and $\sigma$}
In practice, their choice can follow two simple regimes. 
When prior knowledge about the search space is available (e.g., from benchmarks or known ranges of \#Params/FLOPs), $\mu$ and $\sigma$ can be set explicitly from that distribution. 
When prior knowledge is limited or in a search-space agnostic setup, $\mu$ and $\sigma$ can be determined adaptively from the architectures evaluated during the search process itself. 
In this case, their values can be updated using the mean and standard deviation of model sizes observed in the search history, at the end of each search epoch. This online adjustment provides a low-cost and flexible heuristic, ensuring that the regularisation remains aligned with the actual search dynamics.

\subsubsection{Results on the CIFAR-10}
Table \ref{cifar10results} shows the results of architectures found by SWAP-NAS from the DARTS search space for CIFAR-10. The networks' training strategy and hyper-parameters setting exactly follow the setup in DARTS \cite{Ref:10}. Three variations of SWAP-NAS are presented, with different regularisation parameters $\mu$ and $\sigma$. Regardless of these parameters, SWAP-NAS only requires 0.004 GPU days, or 6 minutes. That is 6.5 times faster than the state-of-the-art (TE-NAS). Meanwhile, the architectures found by SWAP-NAS also outperform most previous work. In addition, the capability of model size control is demonstrated. SWAP-NAS-A, with small $\mu$ and $\sigma$ values, generates smaller networks but also suffers a tiny performance deterioration. Conversely, SWAP-NAS-C, with large $\mu$ and $\sigma$, achieves the best error rate but at the cost of a slightly larger network. This capability allows practitioners to find a balance between performance and model size according to the needs of the task. Where SWAP-NAS (Adaptive) reports the results obtained when prior knowledge of the search space is limited. In this setting, $\mu$ and $\sigma$ are adaptively assigned based on the statistical information of previously evaluated architectures, with updates applied during the search process. This heuristic strategy enables end-to-end architecture search in a search-space-agnostic scenario.

\begin{table*}[ht!]
\centering
\caption{Performance comparison between the networks found by SWAP-NAS and other methods on ImageNet. A lower test error rate is better. $\dagger$ means the method adopted DARTS space.}
\begin{tabular}{l|c|c|c|c|c|c}
\hline
  & \textbf{\begin{tabular}[c]{@{}c@{}}Test Error \\ Top1/Top5\end{tabular}} & \textbf{\begin{tabular}[c]{@{}c@{}}Params (M)\end{tabular}} & \textbf{\begin{tabular}[c]{@{}c@{}}GPU Days\end{tabular}} & \textbf{\begin{tabular}[c]{@{}c@{}}Search Method\end{tabular}} & \begin{tabular}[c]{@{}c@{}}{\textbf{Evaluation}}\end{tabular} & \textbf{\begin{tabular}[c]{@{}c@{}}Searched Dataset\end{tabular}} \\ \hline
ProxylessNAS \cite{Ref:15} & 24.9 / 7.5 & 7.1 & 8.3 & Gradient & One-shot & ImageNet \\
DARTS \cite{Ref:10}$\dagger$ & 26.7 / 8.7 & 4.7 & 4 & Gradient & One-shot & CIFAR-10 \\
EvNAS \cite{Ref:62}$\dagger$ & 24.4 / 7.4 & 5.1 & 3.83 & Evolution & One-shot & CIFAR-10 \\
PRE-NAS \cite{Ref:103}$\dagger$ & 24.0 / 7.8 & 6.2 & 0.6 & Evolution & Predictor & CIFAR-10 \\
FairDARTS \cite{Ref:35}$\dagger$ & 26.3 / 8.3 & 2.8 & 0.42 & Gradient & One-shot & CIFAR-10 \\
CARS \cite{Ref:07}$\dagger$ & 24.8 / 7.5 & 5.1 & 0.4 & Evolution & One-shot & CIFAR-10 \\
P-DARTS \cite{Ref:52}$\dagger$  & 24.4 / 7.4 & 4.9 & 0.3 & Gradient & One-shot & CIFAR-10 \\
CTNAS \cite{Ref:72} & 22.7 / 7.5 & - & 0.1+50 & Reinforce & Predictor & ImageNet\\
ZiCo \cite{Ref:119} & 21.9 / - & - & 0.4 & Evolution & Training-free & ImageNet\\
PINAT-T \cite{Ref:120}$\dagger$   & 24.9 / 7.5 & 5.2 & 0.3 & Evolution & Predictor & CIFAR-10 \\
GDAS \cite{Ref:63}   & 27.5 / 9.1 & 4.4 & 0.17 & Gradient & One-shot & CIFAR-10 \\
TE-NAS \cite{Ref:71}$\dagger$ & 24.5 / 7.5 & 5.4 & 0.17 & Pruning & Training-free & ImageNet \\
MeCo \cite{Ref:141}$\dagger$ & 22.2 / - & 7.9 & 0.08 & Pruning & Training-free & CIFAR-10 \\
QE-NAS \cite{Ref:113}$\dagger$ & 25.5 / - & 3.2 & 0.02 & Evolution & Training-free & ImageNet \\
\hline
\textbf{SWAP-NAS ($\mu$=25, $\sigma$=25)} $\dagger$ & 24.0 / 7.6 & 5.8 & 0.006 & Evolution & Training-free & ImageNet   \\
\textbf{SWAP-NAS (Adaptive)} $\dagger$ & 23.3 / 6.8 & 7.3 & 0.006 & Evolution & Training-free & ImageNet   \\
\hline
\end{tabular}

\label{imagenetresults}
\end{table*}

\subsubsection{Results on the ImageNet}
Table \ref{imagenetresults} shows the results on the ImageNet dataset, where the training strategy and hyper-parameters setting are also the same as in DARTS \cite{Ref:10}. The search cost of SWAP-NAS here slightly increased to 0.006 GPU days, or 9 minutes. This is still 2.3 times faster than the state-of-the-art (QE-NAS) yet with better performance.

\subsection{Guidelines for Integrating SWAP with Other Paradigms}
While our experiments demonstrate SWAP-Score within an evolutionary NAS framework, the metric can also complement other paradigms. In gradient-based NAS (e.g., DARTS \cite{Ref:10}, GDAS \cite{Ref:63}), SWAP-Score cannot directly substitute for the gradient signal, but it may serve as a pre-screening mechanism to reduce the candidate set before optimisation. In reinforcement learning NAS (e.g., ENAS \cite{Ref:17}), SWAP-Score can be employed as a reward signal to quickly evaluate sampled policies before committing to training. For predictor-based NAS, SWAP-Score provides a low-cost pseudo-label that can supplement limited ground-truth architecture–accuracy pairs, thereby improving predictor training.

\section{Ablation Studies}
\label{ablation}
A range of ablation studies are conducted to investigate the impacts of batch size, input dimension and input type on the SWAP-Score. Similar to Section \ref{experiment}, both ReLU-based and GELU-based networks are involved in these studies. For ReLU-based networks, 1000 cell networks are sampled from the DARTS architecture space and trained for 200 epochs on the CIFAR-10 dataset. For GELU-based networks, this study reuses the 500 BERT-like Transformers as introduced in Section \ref{sec_metrics_compare_nlp}.

\subsection{Impacts of Batch Sizes}
\label{ablation_batch_size}
\begin{figure}[!h]
  \begin{center}
    \includegraphics[width=0.9\linewidth,height=0.3\textheight]{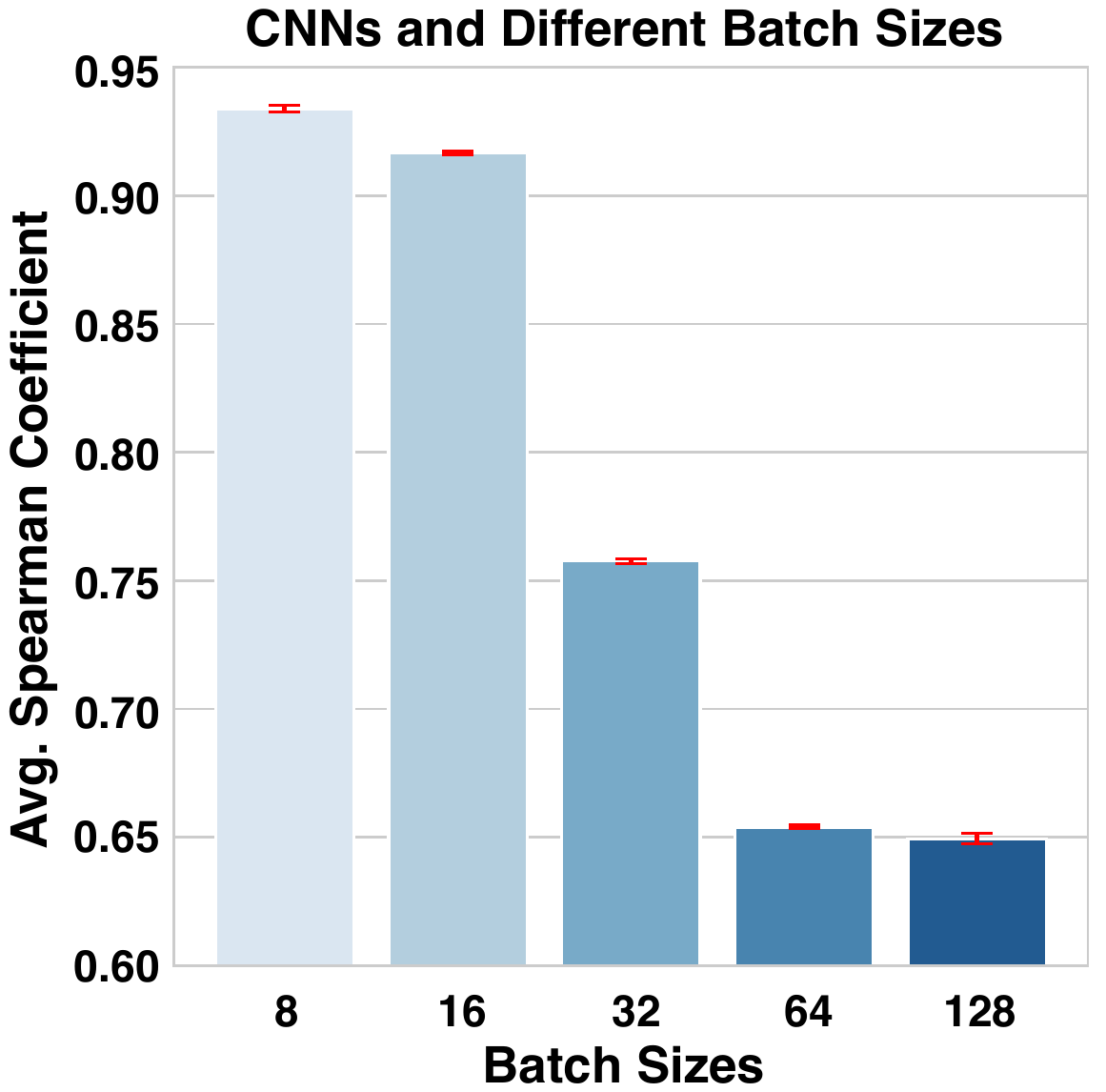}
  \end{center}
  \caption{Spearman's correlation coefficient between SWAP-Score and ground-truth performance of 1000 DARTS CNNs with inputs of different batch sizes. Each setup is repeated 5 times with different random seeds. Each bar shows the average value of 5 runs along with the standard error. The results show a clear tendency for the correlation to decrease as the batch size increase.}
  \label{ablation_batch_size_cnn}
\end{figure}

Figure \ref{ablation_batch_size_cnn} illustrates the average Spearman's correlation coefficient between the SWAP-Score and the ground-truth performance of 1,000 DARTS CNNs on the CIFAR-10 dataset, over five runs with different random seeds. The results demonstrate a clear tendency for the correlation to decrease as the batch size increases, with the Spearman's correlation coefficient initially reaching 0.933 at a batch size of 8 and dropping to 0.649 at a batch size of 128. Notably, the standard error between each run is very small.

\begin{figure}[!h]
  \begin{center}
    \includegraphics[width=0.9\linewidth,height=0.3\textheight]{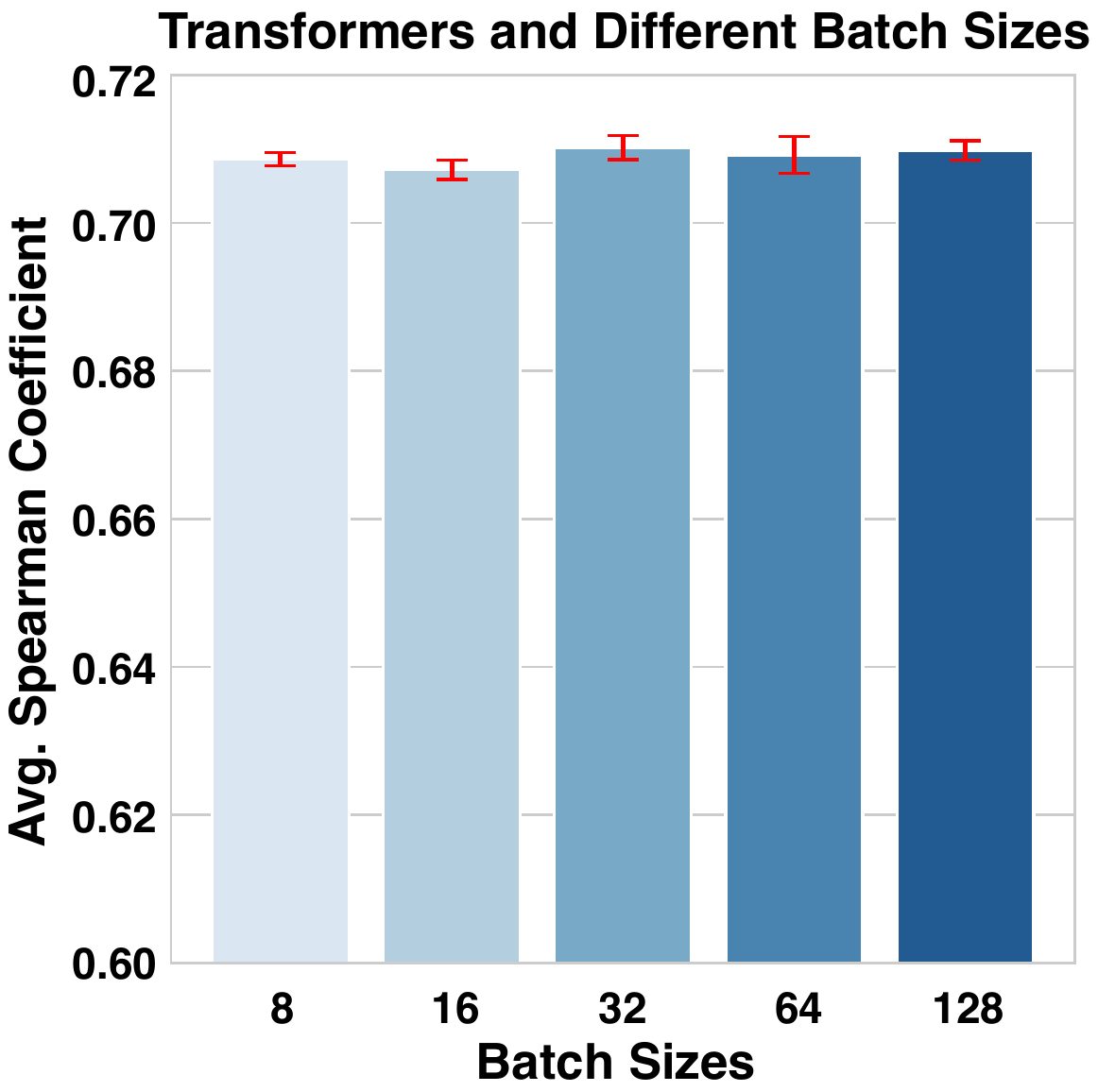}
  \end{center}
  \caption{Spearman's correlation coefficient between SWAP-Score and ground-truth performance of 500 BERT-like Transformers with inputs of different batch sizes. Each setup is repeated 5 times with different random seeds. Each bar shows the average value of 5 runs along with the standard error. The results remain consistent across varying batch sizes.}
  \label{ablation_batch_size_transformer}
\end{figure}

Similarly, Figure \ref{ablation_batch_size_transformer} shows the average Spearman's correlation coefficient between the SWAP-Score and the ground-truth GLUE score of 500 BERT-like Transformers. In contrast to the CNN experiment, the correlation coefficients on the Transformers are not significantly impacted by batch size and remain at approximately $0.71$.

\begin{figure}[!h]
  \begin{center}
    \includegraphics[width=\linewidth,height=0.3\textheight]{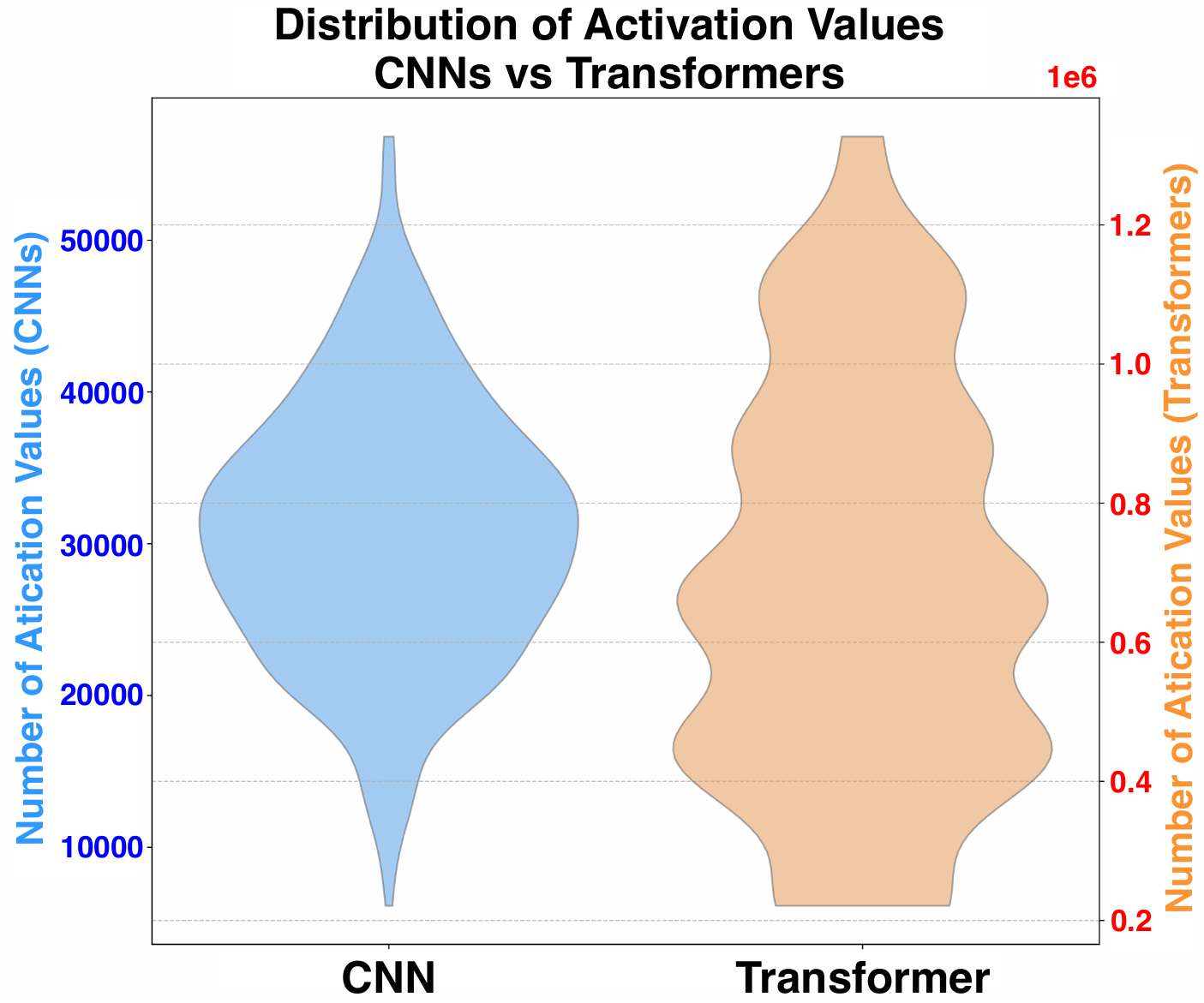}
  \end{center}
  \caption{Distribution of the number of activation values of 1000 DARTS CNNs and 500 BERT-like Transformers. The comparison illustrates that the BERT-like Transformers contain much more activation values than the DARTS CNNs.}
  \label{ablation_num_activation_values}
\end{figure}

To further investigate this phenomenon, we examined the number of activation values of the 1000 DARTS CNNs and the 500 BERT-like Transformers. As shown in Fig. \ref{ablation_num_activation_values},  the number of activation values for the CNNs are distributed in the range of $10^{4}$, while for the Transformers, they are distributed in the range of $10^6$. According to Eq. \ref{swap_set}, the Transformers have much larger $V$ values than the CNNs, making their sample-wise activation patterns more tolerant for larger batch sizes or more input samples, $S$. These results suggest that when calculating SWAP-Score for small neural networks like DARTS CNNs, a smaller batch size yields better performance.

\subsection{Impacts of Input Dimensions and Types}
\label{ablation_dimension_type}

Aside from the batch size of input samples, the impacts of input dimension and input type are also investigated.  We set the batch size to 16 and utilised the 1,000 DARTS CNNs with the CIFAR-10 dataset. For input dimensions, the samples are from the CIFAR-10 dataset, with a size starting from $3\times3\times3$ and gradually increasing to $3\times32\times32$, the original dimension of the CIFAR-10 images. Furthermore, some of the CIFAR-10 images are replaced with Gaussian noise to investigate the impacts of different input types. 

\begin{figure}[!ht]
  \begin{center}
    \includegraphics[width=0.9\linewidth,height=0.3\textheight]{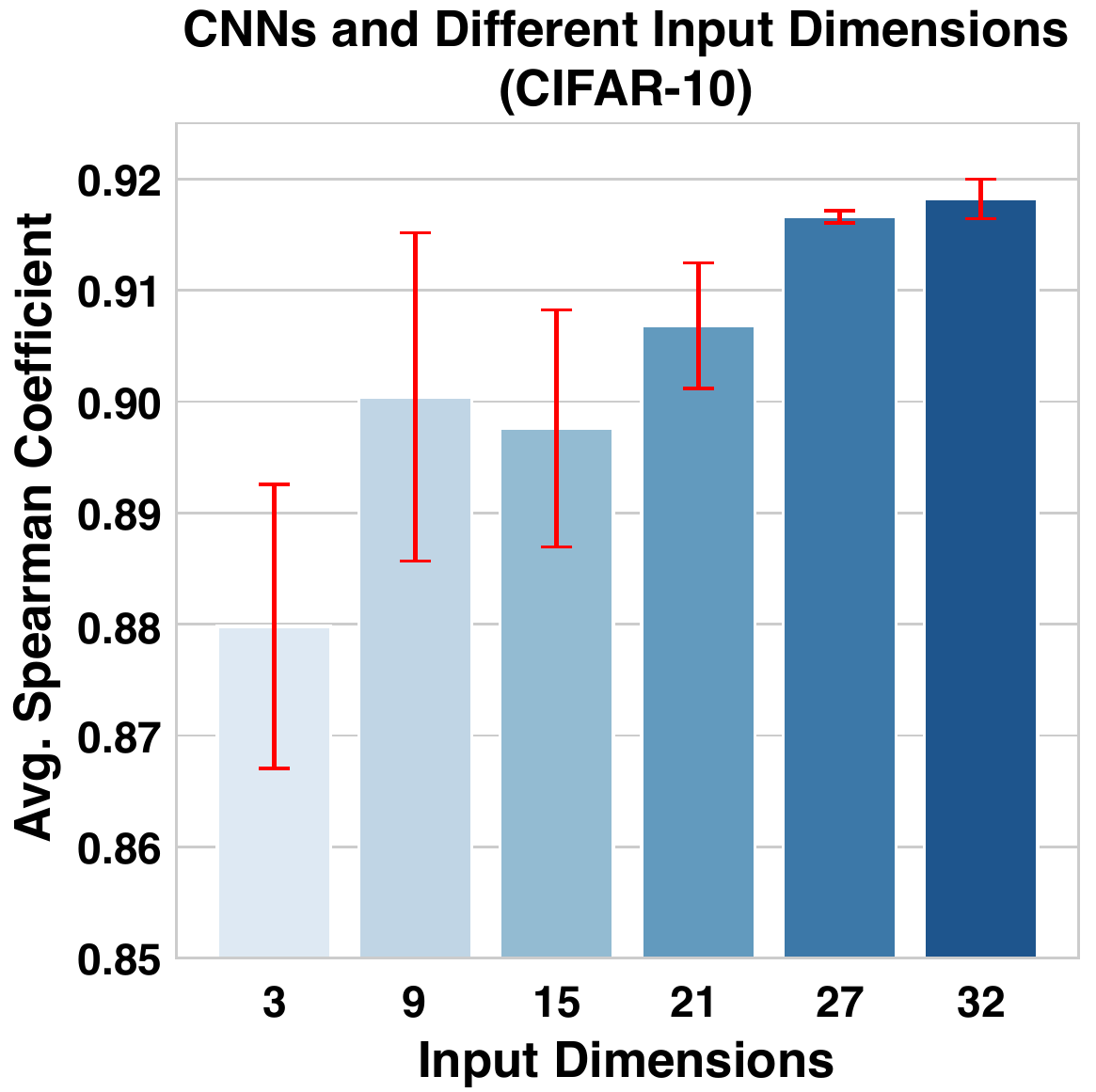}
  \end{center}
  \caption{Spearman's correlation coefficient between SWAP-Score and ground-truth performance of 1000 DARTS CNNs with CIFAR-10 images of different dimensions. Each setup is repeated 5 times with different random seeds. Each bar shows the average value of 5 runs along with the standard error. The results illustrate the sensitivity of SWAP-Score to input image resolution, a larger input dimension yields better correlation.}
  \label{ablation_input_dim_cnn_c10}
\end{figure}

\begin{figure}[!ht]
  \begin{center}
\includegraphics[width=0.9\linewidth,height=0.3\textheight]{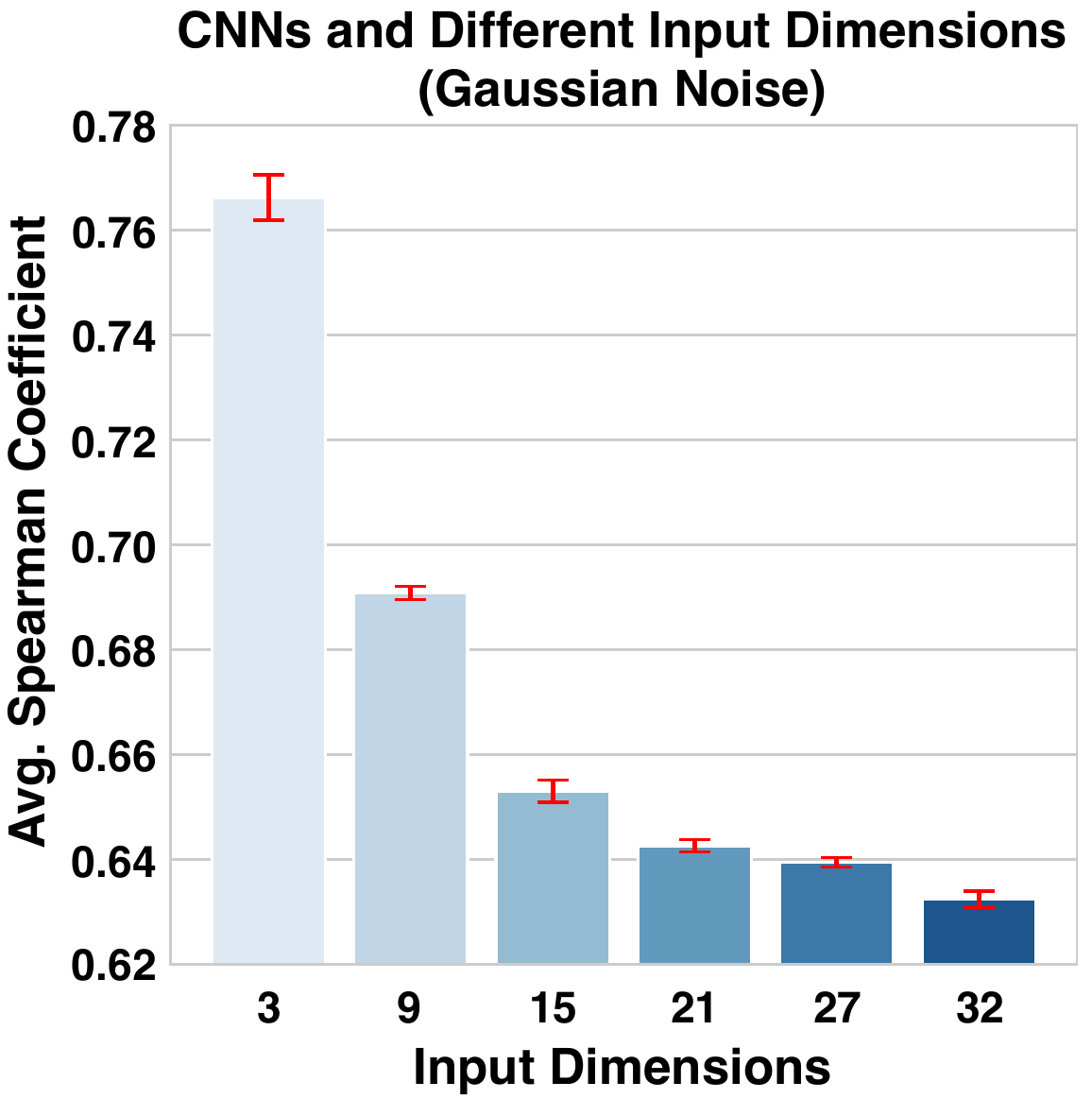}
  \end{center}
  \caption{Spearman's correlation coefficient between SWAP-Score and ground-truth performance of 1000 DARTS CNNs with Gaussian noises of different dimensions. Each setup is repeated 5 times with different random seeds. Each bar shows the average value of 5 runs along with the standard error. Unlike the results in Fig. \ref{ablation_input_dim_cnn_c10}, the Spearman's coefficient decreases as the input dimension increases.}
  \label{ablation_input_dim_cnn_random}
\end{figure}

As shown in Fig. \ref{ablation_input_dim_cnn_c10}, the Spearman's correlation coefficient gradually increases along with the input dimensions. This further verifies our observation in Section \ref{swap} that a larger $V$ offers a higher capacity for distinct patterns, as larger dimensional inputs yield more intermediate values when the architectures of the networks are fixed. In contrast, when the input type is Gaussian noise, the Spearman's correlation coefficient decreases as the input dimension increases, showing in Fig. \ref{ablation_input_dim_cnn_random}. This indicates that original task data offers more informative activation patterns that align with network performance, whereas random noise lacks the structure necessary to provide meaningful signals.

\textbf{Guidelines on Batch Size and Input Dimensions.} Above ablation studies reveal that both batch size and input dimension influence the reliability of SWAP-Score. For CNNs in the DARTS search space, Spearman’s correlation decreases as batch size increases (from 0.933 at $S=8$ to 0.649 at $S=128$). This is because CNNs have a relatively small number of activation values ($V \approx 10^{4}$), so larger batches quickly saturate the diversity of activation patterns. In contrast, BERT-like Transformers ($V \approx 10^{6}$) tolerate larger batch sizes, maintaining stable correlations around 0.71. For small CNNs, use smaller batches (8–16) to preserve sensitivity, while for large-scale models such as Transformers, batch size has little effect and can be set more flexibly.

For input dimensions, higher-resolution task data (e.g., CIFAR-10 at $3\times 32 \times 32$) improves correlation because larger inputs produce more informative intermediate activations. However, when inputs are replaced with Gaussian noise, increasing dimension reduces correlation, as the activations do not reflect meaningful task structure.

\subsection{Effect of Regularisation on Baselines}
\label{ablation_reg}
To further examine whether the gains of Reg.~SWAP arise solely from the Gaussian weighting function $f(\Theta)$, 
we applied the same regularisation to two simple baselines: \#FLOPs and \#Params. Table~\ref{tab_ablation_reg} reports the results across NAS-Bench-101/201/301 architecture spaces.

\begin{table}[h!]
\centering
\caption{Spearman correlation coefficients of regularised and un-regularised baselines. 
We apply the regularisation function $f(\Theta)$ to \#FLOPs and \#Params to test whether the gains of Reg.~SWAP arise solely from regularisation.}
\begin{tabular}{c|c|c|c|c|c}
\hline
              & \begin{tabular}[c]{@{}c@{}}NB101\\ CF10\end{tabular} & \begin{tabular}[c]{@{}c@{}}NB201\\ CF10\end{tabular} & \begin{tabular}[c]{@{}c@{}}NB201\\ CF100\end{tabular} & \begin{tabular}[c]{@{}c@{}}NB201\\ IMGNT\end{tabular} & \begin{tabular}[c]{@{}c@{}}NB301\\ CF10\end{tabular} \\ \hline
\#FLOPs       & 0.37                                                 & 0.69                                                 & 0.71                                                  & 0.67                                                  & 0.43                                                 \\
Reg. \#FLOPs  & 0.76                                                 & 0.69                                                 & 0.71                                                  & 0.67                                                  & 0.53                                                 \\ \hline
\#Params      & 0.38                                                 & 0.72                                                 & 0.73                                                  & 0.69                                                  & 0.46                                                 \\
Reg. \#Params & 0.76                                                 & 0.72                                                 & 0.73                                                  & 0.69                                                  & 0.54                                                 \\ \hline
SWAP          & 0.47                                                 & 0.79                                                 & 0.81                                                  & 0.76                                                  & 0.58                                                 \\
Reg. SWAP     & 0.75                                                 & 0.88                                                 & 0.90                                                  & 0.87                                                  & 0.63                                                 \\ \hline
\end{tabular}
\label{tab_ablation_reg}
\end{table}

We observe that regularisation indeed improves both baselines on certain tasks. For example, on NAS-Bench-101 (CIFAR-10), the Spearman correlation of \#Params increases from $0.38$ to $0.76$ after regularisation, and \#FLOPs rises from $0.37$ to $0.76$. However, on NAS-Bench-201, neither baseline benefits from the regularisation. In contrast, Reg.~SWAP remains consistently superior, with correlations of $0.75$--$0.90$ across all benchmarks, outperforming both Reg.~\#Params and Reg.~\#FLOPs by a clear margin.

These findings demonstrate that while the regularisation function reduces size bias, the activation pattern expressivity captured by SWAP provides additional predictive strength. In other words, Reg.~SWAP benefits from both scale normalisation and structural information, explaining its superior performance over purely regularised size-based baselines.

\section{Conclusion and Future Work}
\label{conclusion}
In this study, we introduced SWAP, Sample-Wise Activation Patterns, and subsequently SWAP-Score, a novel, broadly applicable, and highly effective zero-shot metric based on SWAP. SWAP brings a new perspective for examining the characteristics of networks with different topologies.  It can overcome the limitations of standard activation patterns, offering much better generalisation over various domains and tasks, for both CNN and Transformer architectures.  More importantly, SWAP provides much stronger correlation with the ground-truth performance than existing SOTA zero-shot or training-free metrics across various computer vision and natural language processing tasks. Additionally, the effectiveness of SWAP-Score can be demonstrated through in an important scenario, NAS. When integrated with an evolutionary search algorithm as a new NAS method, SWAP-NAS, a combination of fast low-cost search and highly competitive performance can be achieved on both CIFAR-10 and ImageNet datasets, outperforming state-of-the-art NAS methods. 

In this work, the $Signum$ function is adopted as the indicator function.  In the future work we will see whether it can be replaced by a more fine-grained function, which is capable of capturing nuanced information within the activation values. Moreover, as pre-training has become a standard procedure in modern Large Language Models and Large Vision Models, exploring the impacts of pre-train datasets and hyper-parameters in a zero-shot fashion is also a worthy direction.
 
\bibliographystyle{IEEEtran}
\bibliography{references.bib}

\end{document}